\documentclass[10pt,twocolumn,letterpaper]{article}

\usepackage[pagenumbers]{cvpr} 

\usepackage[dvipsnames]{xcolor}

\usepackage{multirow}
\usepackage{adjustbox}
\usepackage{siunitx}
\usepackage{textcomp}

\usepackage{stmaryrd} 
\usepackage{latexsym} 

\usepackage{url}
\usepackage{kotex} 
\usepackage{booktabs}
\usepackage{xcolor,colortbl}

\newcommand{\figref}[1]{Fig.~\ref{#1}}

\newcommand{\tabref}[1]{Tab.~\ref{#1}}

\newcommand{\secref}[1]{Sec.~\ref{#1}}

\newcommand\blfootnote[1]{%
  \begingroup
  \renewcommand\thefootnote{}\footnote{#1}%
  \addtocounter{footnote}{-1}%
  \endgroup
}
\definecolor{cvprblue}{rgb}{0.21,0.49,0.74}
\usepackage[pagebackref,breaklinks,colorlinks,citecolor=cvprblue]{hyperref}

\def\paperID{14417} 
\def\confName{CVPR}
\def\confYear{2024}

\title{Learning to Control Camera Exposure via Reinforcement Learning}

\author{Kyunghyun Lee*\\
LG AI Research\\
{\tt\small kyunghyun.lee@lgresearch.ai}
\and
Ukcheol Shin*\\
CMU\\
{\tt\small ushin@andrew.cmu.edu}
\and
Byeong-Uk Lee\\
KRAFTON\\
{\tt\small byeonguk.lee@krafton.com}
}

\begin{document}
\twocolumn[{%
\renewcommand\twocolumn[1][]{#1}%
\maketitle
\vspace{-0.3in}
\begin{center}
{
\begin{tabular}{c@{\hskip 0.01\linewidth}c@{\hskip 0.01\linewidth}c}
\multicolumn{3}{c}{\includegraphics[width=0.95\linewidth]{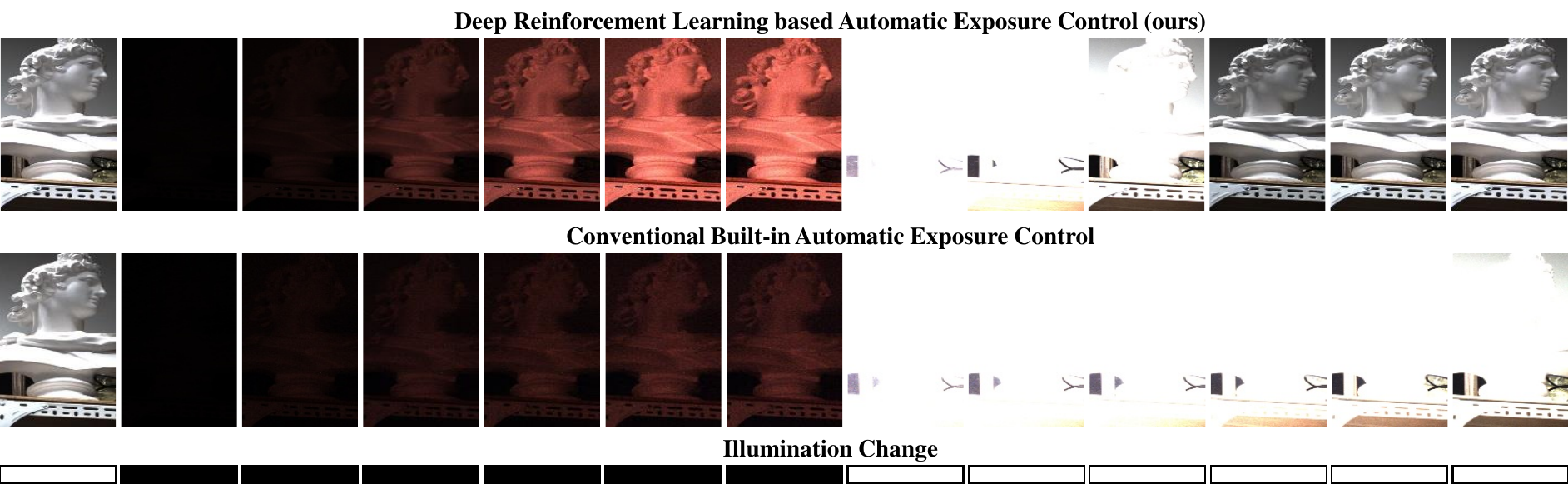}} \\ 
\multicolumn{3}{c}{\footnotesize \textbf{(a) Automatic exposure control for sudden lighting changes}} \\
\includegraphics[height=0.15\linewidth]{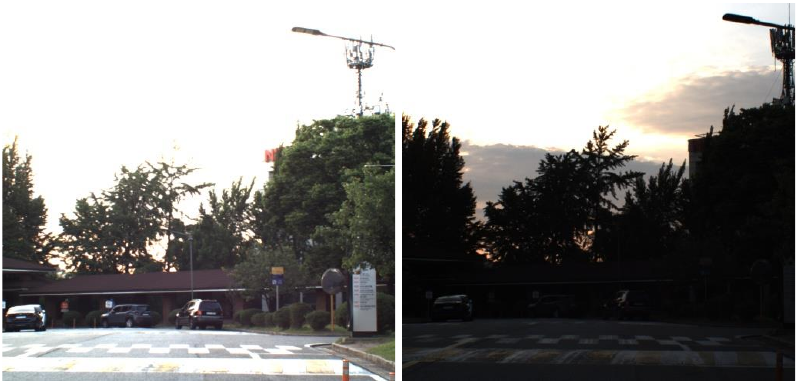} &
\includegraphics[height=0.15\linewidth]{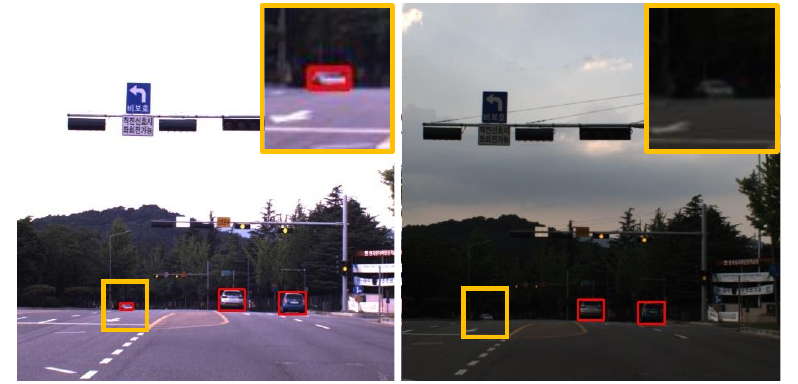} &
\includegraphics[height=0.15\linewidth]{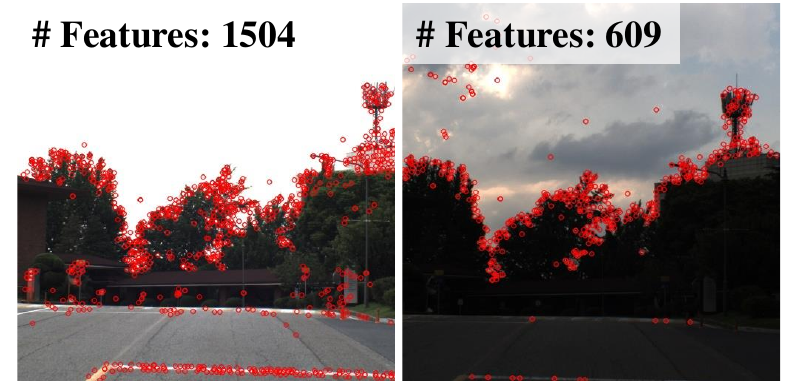} \\ 
\textbf{\footnotesize (1) Well-exposed image acquisition} & 
\textbf{\footnotesize (2) Object detection} & 
\textbf{\footnotesize (3) Feature extraction} \\
\multicolumn{3}{c}{\footnotesize \textbf{(b) Effectiveness on various vision applications (left: ours, right: built-in AE)}  } \\
\end{tabular}
}
\captionof{figure}{
{\bf Automatic camera exposure control via deep reinforcement learning.} 
Our proposed method, named DRL-AE, trains an agent to control camera exposure parameters (\ie, exposure time and gain) to acquire well-exposed images with rapid convergence and real-time processing (1\textit{ms} on a CPU device).
The trained agent instantly converges within five frames under dramatic lighting change scenario (a) and affects the performance of various vision applications (b), compared to the camera built-in AE controller~\cite{muramatsu1997photometry,sampat1999system}. 
}
\vspace{0.13in}
\end{center}
}]

\begin{abstract}
Adjusting camera exposure in arbitrary lighting conditions is the first step to ensure the functionality of computer vision applications\blfootnote{*Both authors contributed equally to this work.}. 
Poorly adjusted camera exposure often leads to critical failure and performance degradation.
Traditional camera exposure control methods require multiple convergence steps and time-consuming processes, making them unsuitable for dynamic lighting conditions.
In this paper, we propose a new camera exposure control framework that rapidly controls camera exposure while performing real-time processing by exploiting deep reinforcement learning.
The proposed framework consists of four contributions: 
1) a simplified training ground to simulate real-world's diverse and dynamic lighting changes, 
2) flickering and image attribute-aware reward design, along with lightweight state design for real-time processing, 
3) a static-to-dynamic lighting curriculum to gradually improve the agent's exposure-adjusting capability, and 
4) domain randomization techniques to alleviate the limitation of the training ground and achieve seamless generalization in the wild.
As a result, our proposed method rapidly reaches a desired exposure level within five steps with real-time processing (1\textit{ms}).
Also, the acquired images are well-exposed and show superiority in various computer vision tasks, such as feature extraction and object detection.
\end{abstract}

\vspace{-0.3in}
\section{Introduction}

\quad Camera exposure control is the task of adjusting exposure level by controlling exposure time, gain, and aperture to achieve a desired level of brightness and image quality for a given scene.
Poorly adjusted exposure parameters result in over-exposed, under-exposed, blurry, or noisy images, which can cause performance degradation in image-based applications and, in the worst cases, even life-threatening accidents.
Therefore, finding proper camera exposure is the first primary step to ensure the functionality of computer vision applications, such as object detection~\cite{redmon2017yolo9000,he2017mask}, semantic segmentation \cite{ronneberger2015u,kirillov2018panoptic}, depth estimation \cite{zhao2020monocular,lee2021depth}, and visual odometry \cite{forster2014svo,mur2017orb}. 

There are several essential requirements in camera exposure control. 
The rapid convergence must be guaranteed to maintain an appropriate exposure level under dynamic light-changing scenarios. 
Also, the exposure control loop is one of the lowest loops in the camera system.
Therefore, lightweight algorithm design must be considered for on-board level operation. 
Finally, the quality of a converged image should not be sacrificed to meet the requirements. 

Further, the number of simultaneously controlled parameters is also important because it affects the converge time and final quality of the converged image.
One-by-one control methods~\cite{muramatsu1997photometry,sampat1999system,shim2018gradient} control exposure parameters in a one-by-one manner to achieve a desired exposure level, rather than joint controlling exposure parameters.
However, the converged parameters are often not optimal, such as [long exposure time, low gain] and [short exposure time, high gain] pairs. 
As a result, the values result in undesirable image artifacts, such as motion blur due to long exposure time or severe noise due to high gain.

Joint exposure parameter control~\cite{kim2018exposure,shin2019camera,kim2020proactive,tomasi2021learned,wang2022automated} often needs multiple searching steps in a wide range of searching space to find an optimal combination.
As a result, they cause a flickering effect and slow convergence speed.
Also, the recent methods require high-level computational complexity due to its optimization algorithm~\cite{kim2018exposure,kim2020proactive}, image assessment metric~\cite{shim2018gradient,shin2019camera,kim2018exposure,kim2020proactive}, and GPU inference~\cite{tomasi2021learned}.

In this paper, we propose a new joint exposure parameter control method that exploits reinforcement learning to achieve instant convergence and real-time processing. 
The proposed framework consists of four contributions: 
\begin{itemize}
    \item A simplified training ground to simulate real-world's diverse and dynamic lighting changes.
    \item Flickering and image attribute-aware reward design, along with lightweight and intuitive state design for real-time processing.
    \item A static-to-dynamic lighting curriculum learning to gradually improve agent's exposure adjusting capability.
    \item Domain randomization techniques to alleviate the limitation of the training ground and achieve seamless generalization in the wild without additional training.
\end{itemize}

The proposed method is thoroughly validated in three different environments: light-controlled darkroom, exposure control dataset~\cite{shin2019camera}, and real-world environments.
We demonstrate that our proposed method rapidly adjusts camera exposure within five steps with real-time processing of 1 \textit{ms}.
Also, the images acquired from our method are well-exposed and show superiority in numerous computer vision tasks, such as feature extraction and object detection.

\section{Related Work}
\begin{figure*}[t]
\begin{center}
\includegraphics[width=0.95\linewidth]{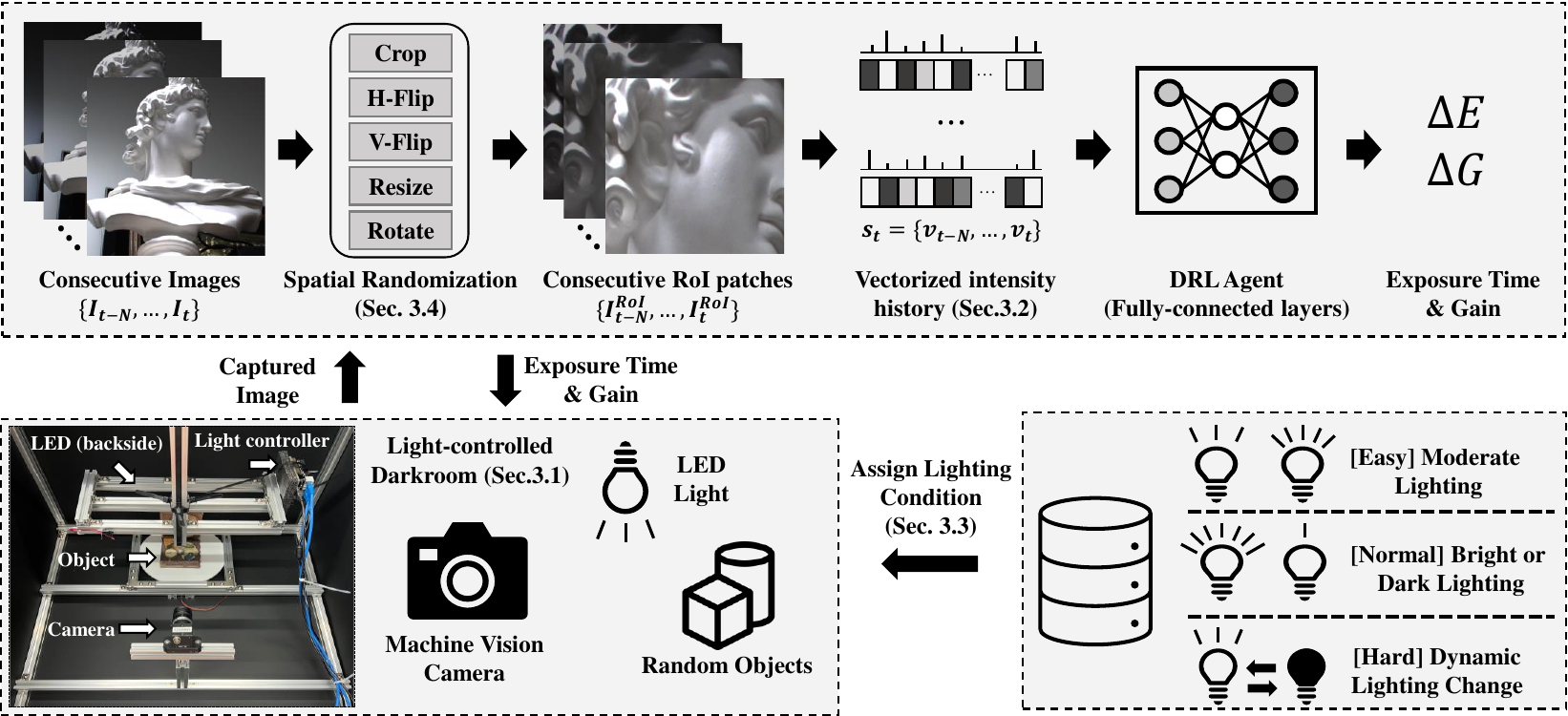}
\caption{
{\bf Training framework overview.} Our DRL agent is trained with the SAC algorithm in the light-controlled dark room environment. For each episode, a lighting condition is assigned by the current curriculum level. 
The lighting condition can be fixed at random brightness or dynamically changed within each episode, depending on the level.
Given the lighting condition, the agent takes a vectorized intensity history for a randomly selected RoI patch as a state. 
Afterward, the agent estimates exposure time and gain differences that maximize a reward function. 
With this framework, the trained agent successfully generalized into a real environment without additional training.}
\label{fig:system_overview}
\end{center}
\vspace{-0.3in}
\end{figure*}

\subsection{Optimization-based Exposure Control}

\quad One branch to control camera exposure parameters is exploiting white-box and black-box optimizations to find optimal parameters for the desired exposure level.
Camera built-in Auto-Exposure (AE) control methods~\cite{muramatsu1997photometry,sampat1999system} adjust exposure parameters (\ie, exposure time and gain) based on differentiable optimization by using the equation between Exposure Value (EV) and exposure parameters~\cite{ray2000manual}.
They control exposure parameters one-by-one to achieve pre-defined image brightness. 
These built-in AE methods provide real-time processing ability but result in non-optimum solutions (\eg, long exposure time and low gain) and limited scalability.
The former limitation causes motion blur and severe image noise due to long exposure time and high gain.
The latter limitation indicates the methods cannot be extendable to maximize other image attributes, such as image gradient or entropy.

Recent AE algorithms are designed to maximize desirable image attributes for computer vision applications, such as image gradient~\cite{shim2018gradient,zhang2017active,shin2022drl}, entropy~\cite{kim2018exposure,kim2020proactive}, noise level~\cite{shin2019camera}, and optical flow~\cite{han2023camera}.
However, these algorithms mainly focused on image metrics for better quality, not the control method. 
Therefore, they adopt heuristic control algorithm~\cite{shim2018gradient} or black-box optimization methods, such as Bayesian optimization~\cite{kim2018exposure,kim2020proactive} and Nelder-Mead optimization~\cite{shin2019camera}.
These black-box optimizations and attribute assessment metrics often require multiple explorations that cause a flickering effect, multiple steps to converge, or heavy computation time.
Differing from the previous method, the proposed method provides rapid convergence, real-time processing, and potential scalability by exploiting Deep Reinforcement Learning (DRL).

\subsection{Data-driven Exposure Control}

\quad Another emerging branch of AE is utilizing a neural network to predict appropriate exposure parameters.
\citet{tomasi2021learned} proposed an exposure parameter estimation network that predicts optimal exposure time and gain for each given image.
The neural network, consisting of a few convolution and linear layers, is trained with Ground Truth (GT) exposure parameters in a supervised manner.
However, the GT label generation needs a time-consuming and complicated process that collects multiple images with varying exposure parameters for every scene.
Also, a heavy computation caused by the use of convolution layers is another drawback.
Differing from the method, our method trains a neural network by maximizing a reward function without relying on specific GT data or generation processes.

\section{DRL for Automatic Exposure Control}
\label{sec:method}

\quad Applying DRL to Automatic Exposure (AE) control task while achieving rapid convergence speed and real-time processing presents several challenges.
Our proposed method provides effective solutions for the following questions.
\begin{enumerate}
    \item Environment: what is the most effective form of training environment to learn camera exposure control?
    \item Reward: what aspects does the agent need to maximize?
    \item State: where is the bottleneck for real-time processing?
    \item Generalization: how to achieve seamless generalization in the wild lighting condition?
\end{enumerate}

\subsection{Training Environment}
\label{subsec:env}
\quad DRL requires a large number of samples and interaction with the environment to train the agent.
Also, the environment needs to provide a diverse and wide range of problems for the agent to solve.
The optimal form of exposure control is to instantly adjust exposure parameters for a variety of lighting conditions, from static lighting conditions to dramatic lighting changes.
To this end, the training environment for camera exposure control must provide diverse lighting change scenarios to the agent.

Numerous options exist, such as a simulation, a real-world environment with natural sunlight, and a controlled real-world environment.
The simulation can provide various images and lighting conditions but has the limitations of an imperfect exposure parameter implementation and a sim-to-real domain gap. 
On the other hand, the real-world environment with natural sunlight has no domain gap, but the lighting conditions change very slowly. 
Therefore, we construct a controlled real-world environment in a darkroom with controllable LEDs to adjust lighting conditions.

The constructed light-controlled darkroom is shown in~\figref{fig:system_overview}. 
The environment has one machine vision camera, a random target object, a light controller, and an LED bar.
The environment provides a random lighting scenario from dark to bright light conditions, and the agent adjusts exposure parameters to capture a high-quality image of the target object in the suggested scenario. 
The detailed sensor specification can be found in the supplementary material.

\subsection{State, Action, and Reward Design}
\label{subsec:state}
\subsubsection{State Design: Vectorized Intensity History}
\quad The widely adopted state design in related fields is utilizing a feature map from a pre-trained network~\cite{tomasi2021learned,shin2022drl}.
However, we found the CNN feature is not effective in the exposure control task due to its disadvantages: 1) unclear relation between CNN feature and exposure level, 2) generalization problem of CNN feature, and 3) heavy computation for on-board devices due to multiple convolution layers. 

Therefore, instead of using the CNN feature or other complicated features as a state, we utilize vectorized image intensity as a straightforward and lightweight state representation for the exposure control task. 
As shown in~\figref{fig:system_overview}, we first vectorize image intensity map from a Region-of-Interest (RoI) patch of gray-scale image $\mathbb{R}^{H_{RoI}\times W_{RoI}}$ to the intensity vector $\mathbb{R}^{S}$. 
The RoI patch can be an entire image area or specific regions decided by a domain randomization process. 
After that, we stack consecutive frame's intensity vectors to embed previous state history, as follows: 
\begin{align*}
    v_t &= f(I_t^{RoI}), \quad v_t \in \mathbb{R}^{S}, \\
    s_t &= \text{concat}(v_{t-n}, ..., v_{t-1}, v_t),
\end{align*}
where $I_t$ is a normalized image at time step $t$, $f(\cdot)$ is an averaging process through the x-axis of a gray-scale image, defined as $\frac{1}{H_{RoI}}\sum_x\text{gray}(I_t^{RoI})$, $v_t$ represents the vectorized intensity that has a dimension of $S$, and $s_t$ is a state vector. 
To ensure a fixed and reasonable state length, we resize the given RoI patch to have $S=128$ and stack previous states with $n=3$.
The proposed state design is effective, straightforward, and computationally efficient compared to the CNN feature, as described in~\secref{subsec:exp1}.

\subsubsection{Action Design: Relative and Continuous Action}
\quad Obviously, we have two controllable parameters, exposure time and gain, to adjust camera exposure. 
However, there are numerous options for its action design, such as 1) discrete vs. continuous action space, and 2) absolute vs. relative action range.
A discrete action space discretizes the action range into a few action values.  
It has the advantage that the training process is simplified, but there is an approximation gap to the optimal values. 
The absolute and relative action range is about the change of camera parameters. 
In the absolute action range, an action value is directly matched to the specific value of camera parameters. 
On the other hand, in the relative action range, an action value indicates the amount of change in camera parameters. 

Among the options, we select continuous-relative action space. 
This is because our goal is rapid convergence with minimum exploration step, but discrete action space needs multiple steps and often does not converge, depending on its quantization level. 
Also, we empirically found that the absolute action range often induces a flickering effect and unstable convergence, as described in~\secref{subsec:exp1}. 

\subsubsection{Reward Design: Flickering and Image Attribute}
\quad The desirable behavior of exposure control is maximizing image attributes, such as sharp edge, moderate brightness, and low-level noise, and maintaining image attributes during exposure parameter transition.
Therefore, we designed the reward function from three perspectives: 1) a moderate brightness level to provide clear visibility and edge information, 2) a smoothed exposure transition to ensure stable convergence and prevent flickering, and 3) a low-level noise to provide clear image and avoid too-high gain value. 
The designed reward functions are as follows:
\begin{align*}
    \mathcal{R}_{mean} &= \textstyle\frac{1}{P}\textstyle\sum_{xy}|I_t^{RoI} - M|^{p_m}, \\
    \mathcal{R}_{flk} &= \textstyle\frac{1}{P}\textstyle\sum_{xy}||I_{t}^{RoI} - I_{t-1}^{RoI}||, \\
    \mathcal{R}_{noise} &= \textstyle\frac{1}{P}\textstyle\sum_{xy} sobel(I_{t}^{RoI}), \\
    \mathcal{R}_{total} &= w_m\mathcal{R}_{mean} + w_f\mathcal{R}_{flk} + w_n\mathcal{R}_{noise},
\end{align*}
where $P$ is the number of pixels, $M=0.5$ indicates mid-tone brightness, $p_m=0.5$ is a parameter for non-linearity and $sobel$ is a gradient operator.
Also, we set $w_m=1.5$, $w_f=-1.0$, $w_n=-0.1$ in practice. 
The proposed reward design might be a primitive and basic form for camera exposure control, however, it can be easily extendable by incorporating modern image assessment metrics~\cite{shin2019camera,kim2020proactive,han2023camera}.

\subsection{Static-to-dynamic Lighting Curriculum}
\label{subsec:curriculum}
\quad In the wild, the agent must be able to control exposure parameters for a variety of lighting change scenarios.
However, training every scenario simultaneously results in an unstable training process and poor generalization.
Therefore, we propose static-to-dynamic curriculum strategy that starts with a simple control task and gradually experiences dynamic and dramatic lighting change scenarios. 
In the end, the trained models possess a comprehensive exposure control capability for diverse lighting conditions.

We divide the difficulty of lighting conditions into three levels: easy, normal, and hard.
The easy level has static lighting conditions with moderate brightness.
The normal level also has a fixed brightness but with a darker or brighter than easy level.
Lastly, in the hard level, the LED brightness dynamically changes from dark to bright or the opposite way during each scenario.
The probability of each level is gradually updated according to the proceeded training episode $t_e$.
The probability set $[p_{e}, p_{n}, p_{h}]$ starts from $[1,0,0]$, through $[0,1,0]$, and ends with [$p_{e}^f, p_{n}^f, p_{h}^f] = [0.1, 0.4, 0.5]$.
In summary, the probability for each difficulty level is updated as follows:
\begin{align*}
p_{e} &= \begin{cases}
1, & t_e < T_{e} \\
\frac{(t_e - T_{e})}{(T_{n} - T_{e})}, & T_{e} \leq t_e < T_{n} \\
p_{e}^f, & T_{n} \leq t_e
\end{cases} \\
p_{n} &= \begin{cases}
0, & t_e < T_{e} \\
1-\frac{(t_e - T_{e})}{(T_{n} - T_{e})}, & T_{e} \leq t_e < T_{n} \\
p_{n}^f, & T_{n} \leq t_e
\end{cases} \\
p_{h} &= \begin{cases}
0, & t_e < T_{n} \\
p_{h}^f, & T_{n} \leq t_e
\end{cases}
\end{align*}
We use $T_e=25,000, T_n=45,000$ in practice. 

\subsection{Spatial Domain Randomization}
\label{subsec:domain_rand}
\quad In the wild, the agent encounters various surrounding environments and object contexts, such as office, road, tunnel, and mountain.
Although the light-controlled darkroom can provide various lighting scenarios, it is difficult to contain diverse environments and contexts because it only has a few target objects with a fixed background. 
Therefore, without proper randomization techniques, the agent may overfit to perform exposure control for only a few target objects, resulting in generalization failure in the wild.

The main idea is to provide as much diverse image structure and context information as possible by augmenting the image from the darkroom environment.
Specifically, we spatially augment the images with random flipping, cropping, rotating, and resizing but do not change color and brightness information. 
Each augmentation and its parameter is randomly selected at the beginning of each training episode and fixed during the episode.
With the proposed domain randomization technique, the trained agent can be generalized in the real world without any fine-tuning. 

\subsection{Policy Optimization}
\quad As our action space is continuous, we use the SAC \cite{haarnoja2018soft} algorithm. 
We excluded on-policy algorithms like PPO \cite{schulman2017proximal} because they are widely known to be less sample efficient than the off-policy algorithms like SAC, TD3, and DDPG.
We tested TD3 \cite{fujimoto2018addressing} and DDPG \cite{lillicrap2015continuous} as well, but SAC showed the best result. 

SAC algorithm is a kind of actor-critic algorithm, which has critic $Q(\theta)$ and actor $\pi(\phi)$. 
The objective functions to update the critic are as follows:
\begin{align*}
& J_Q(\theta) = \mathbb{E}_{(s_t, a_t)\sim \mathcal{D}} \left[ \frac{1}{2}\left( 
Q_\theta(s_t, a_t) - \hat{Q}(s_t, a_t) \right)^2\right], \\
& \hat{Q}(s_t, a_t) = r(s_t, a_t) + \gamma \mathbb{E}_{s_{t+1}\sim p}[V_{\bar{\theta}}(s_{t+1})],
\end{align*}
where $\mathcal{D}$ is a replay buffer, $r(s_t, a_t)$ is a reward function and $\gamma$ is a discount factor.
The objective functions for updating the actor is as follows:
\begin{align*}
    J_\pi(\phi) = \mathbb{E}_{s_t \sim \mathcal{D}}[\mathbb{E}_{a_t\sim\pi_{\phi}}[\alpha \text{log}(\pi_\phi(a_t|s_t))-Q_\theta(s_t, a_t)] ],
\end{align*}
with $\alpha$ defined as a temperature parameter. 

\section{Experiments}
In this section, we validate our proposed method in three different environments: light-controlled darkroom, exposure control dataset~\cite{shin2019camera}, and real-world environments.
Throughout the experiments, we provide the validation result of DRL design components, ablation study of reward and training strategy, convergent step comparison, comparison with built-in AE for object detection and feature extraction, and computational time analysis.

\begin{table}[t]
\caption{\textbf{Self-evaluation of DRL-AE framework in the light-controlled darkroom.} DR and CL indicate domain randomization and curriculum learning. "-" indicates the agent doesn't converge. The best performance in each block is highlighted in \textbf{bold}. 
}
\label{tbl:rlparameter}

\begin{center}
\begin{small}
\vskip -0.2in
\begin{tabular}{c|cc|cc}
\toprule
Framework & \multicolumn{2}{c|}{\multirow{2}{*}{Methods}} & Reward & Frames to \\
Component & & & per Frame & Converge \\
\midrule
 \multirow{3}{*}{\shortstack{RL \\ Algorithm}} & \multicolumn{2}{c|}{DDPG~\cite{lillicrap2015continuous}} & 1.11 & - \\
                                               & \multicolumn{2}{c|}{TD3~\cite{fujimoto2018addressing}}  & 1.03 & - \\
                                               & \multicolumn{2}{c|}{SAC~\cite{haarnoja2018soft}}  & \textbf{1.61} & \textbf{5} \\
 \midrule
 \multirow{2}{*}{\shortstack{State}} & \multicolumn{2}{c|}{CNN}     & 0.85 & - \\
                                               & \multicolumn{2}{c|}{Vector}  & \textbf{1.61} & \textbf{5} \\
 \midrule
 \multirow{2}{*}{\shortstack{Action}} & \multicolumn{2}{c|}{Absolute}  & 0.65 & - \\
                                                 & \multicolumn{2}{c|}{Relative}  & \textbf{1.61} & \textbf{5} \\
 \midrule
 
 \multirow{5}{*}{\shortstack{Reward}} & $R_{flk}$ & $R_{noise}$ &  &  \\
 \cline{2-3}
 & &  & 1.41 & -  \\
 & \checkmark &  & 1.44 & 16  \\
 &  & \checkmark & 1.35 & 9  \\
 &  \checkmark & \checkmark & \textbf{1.61} & \textbf{5}  \\
 \midrule
 \multirow{4}{*}{\shortstack{Training \\ Strategy}} & DR & CL &  &   \\
 \cline{2-3}
 & \checkmark &  & - & -  \\
 &  & \checkmark & 1.64 & 15  \\
 &  \checkmark & \checkmark & \textbf{1.61} & \textbf{5} \\

\bottomrule
\end{tabular}

\end{small}
\end{center}
\vskip -0.3in
\end{table}

\subsection{Self-evaluation in Light-controlled Darkroom}
\label{subsec:exp1}

\quad 
We first validate our DRL design components and their variants in the light-controlled darkroom.
We utilize reward per frame and the averaged number of frames to converge as evaluation metrics to measure image quality and convergence speed, respectively.
Here, when the difference between current and previous images is less than a certain threshold, we regard it as the convergence.
The testing scenarios include various lighting conditions, such as fixed lighting, progressive light changes, and dynamic light changes.
The results are shown in \tabref{tbl:rlparameter}.

We found SAC method~\cite{haarnoja2018soft} shows the best result among the off-policy RL methods.
Other algorithms can reach up to 1.0 reward per frame, but they usually do not converge well by showing oscillation.
For the state design, the CNN feature is not desirable for the exposure control task due to its intensity-agnostic property.
Also, absolute action space seems to make the overall learning process difficult because it needs to estimate the optimum values directly.

The reward function and the training strategies play an important role in the stable and rapid convergence process and image attribute preservation. 
$R_{noise}$ suppress high noise level and regularize gain parameter control, leading to better convergence. 
Also, $R_{flk}$ makes the agent preserve the image attribute during the exposure transition.
CL makes the agent encompass the comprehensive exposure control capability for the test set's various lighting conditions. Additionally, DR allows the agent to quickly converge for arbitrary context by increasing generalization ability.

\begin{figure}[t]
\begin{center}
\begin{tabular}{c@{\hskip 0.005\linewidth}c@{\hskip 0.005\linewidth}c@{\hskip 0.005\linewidth}c@{\hskip 0.005\linewidth}c}
\multicolumn{5}{c}{\footnotesize Shin \etal \cite{shin2019camera}}\\
\includegraphics[width=0.18\linewidth]{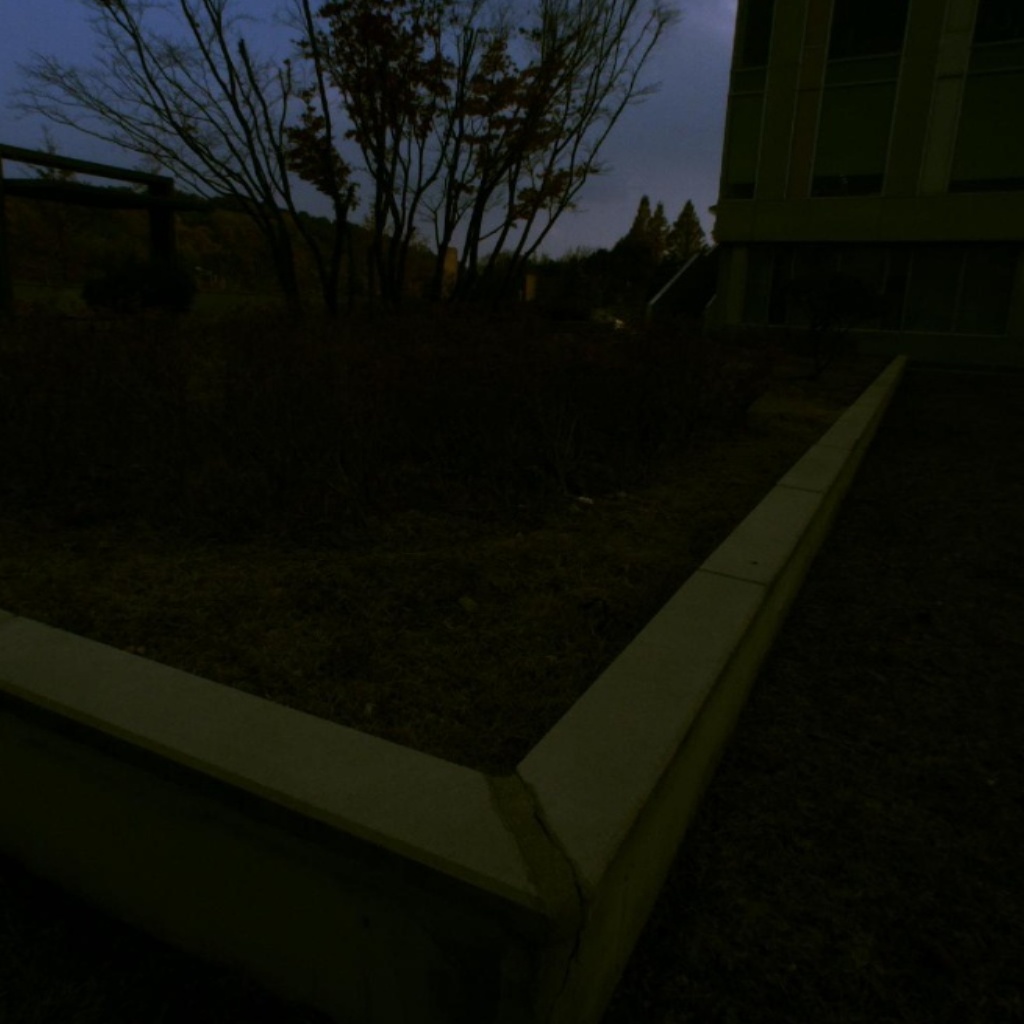} &
\includegraphics[width=0.18\linewidth]{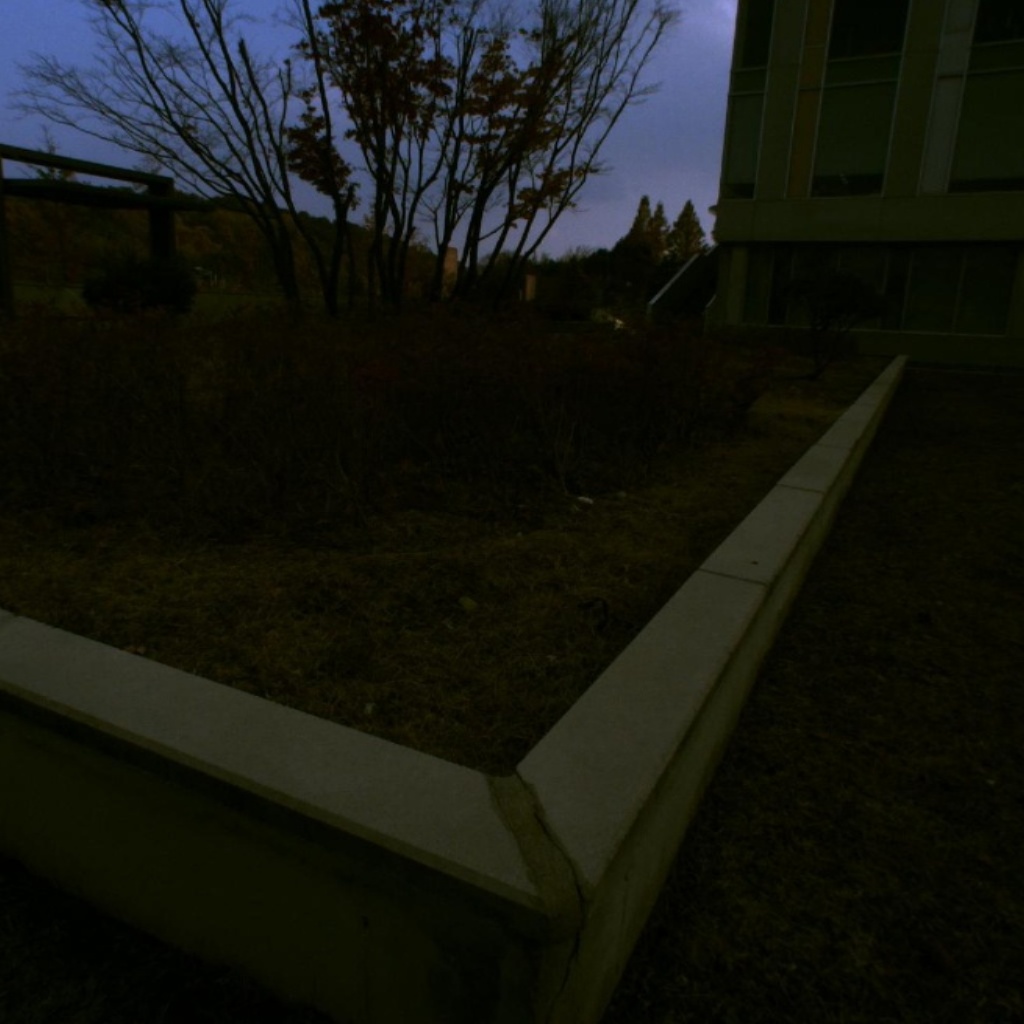} &
\includegraphics[width=0.18\linewidth]{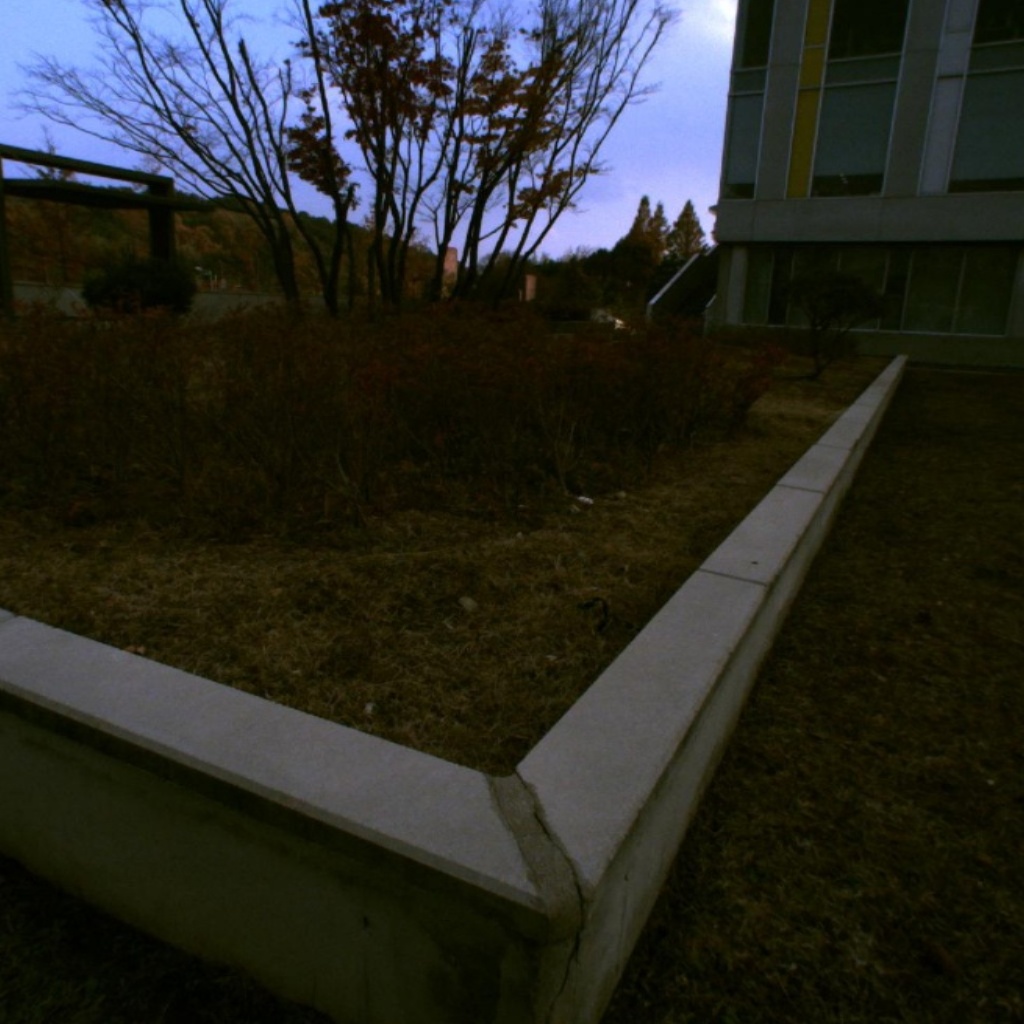} &
\includegraphics[width=0.18\linewidth]{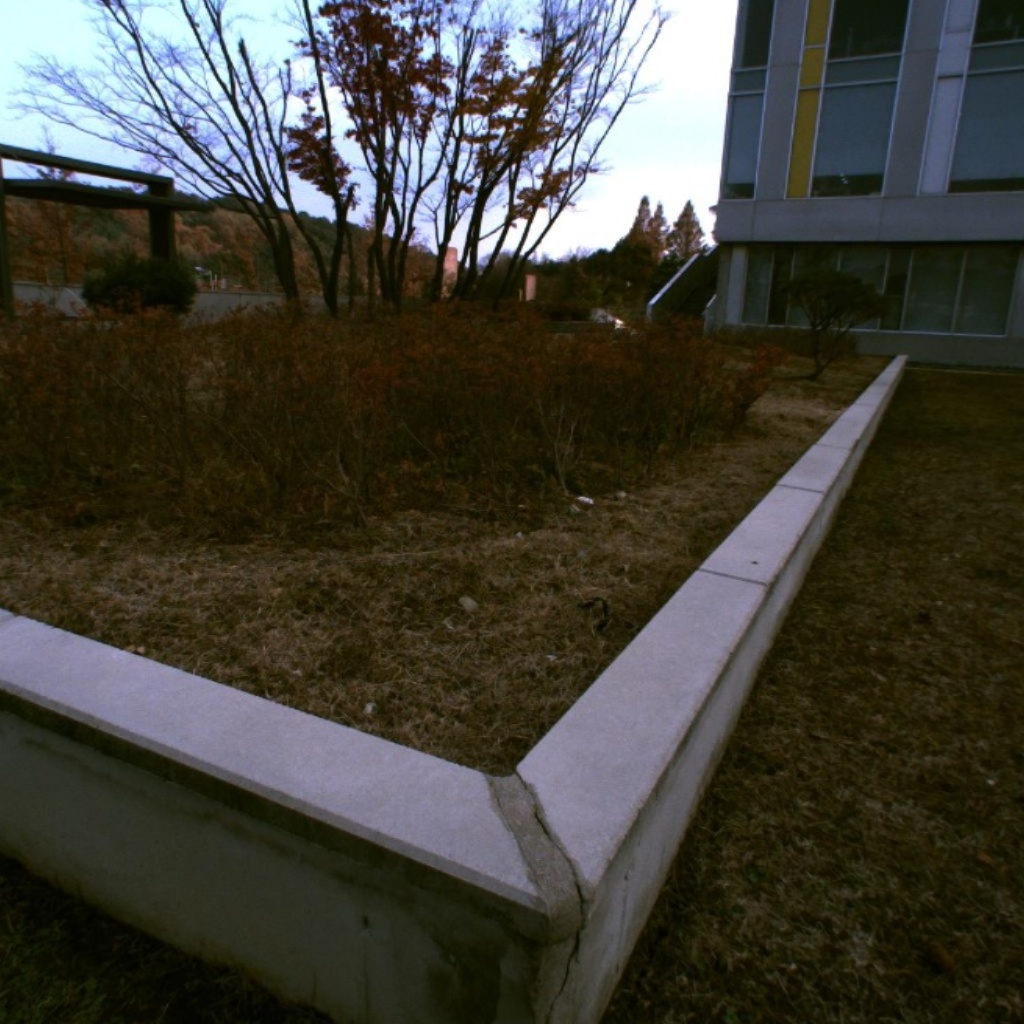} &
\includegraphics[width=0.18\linewidth]{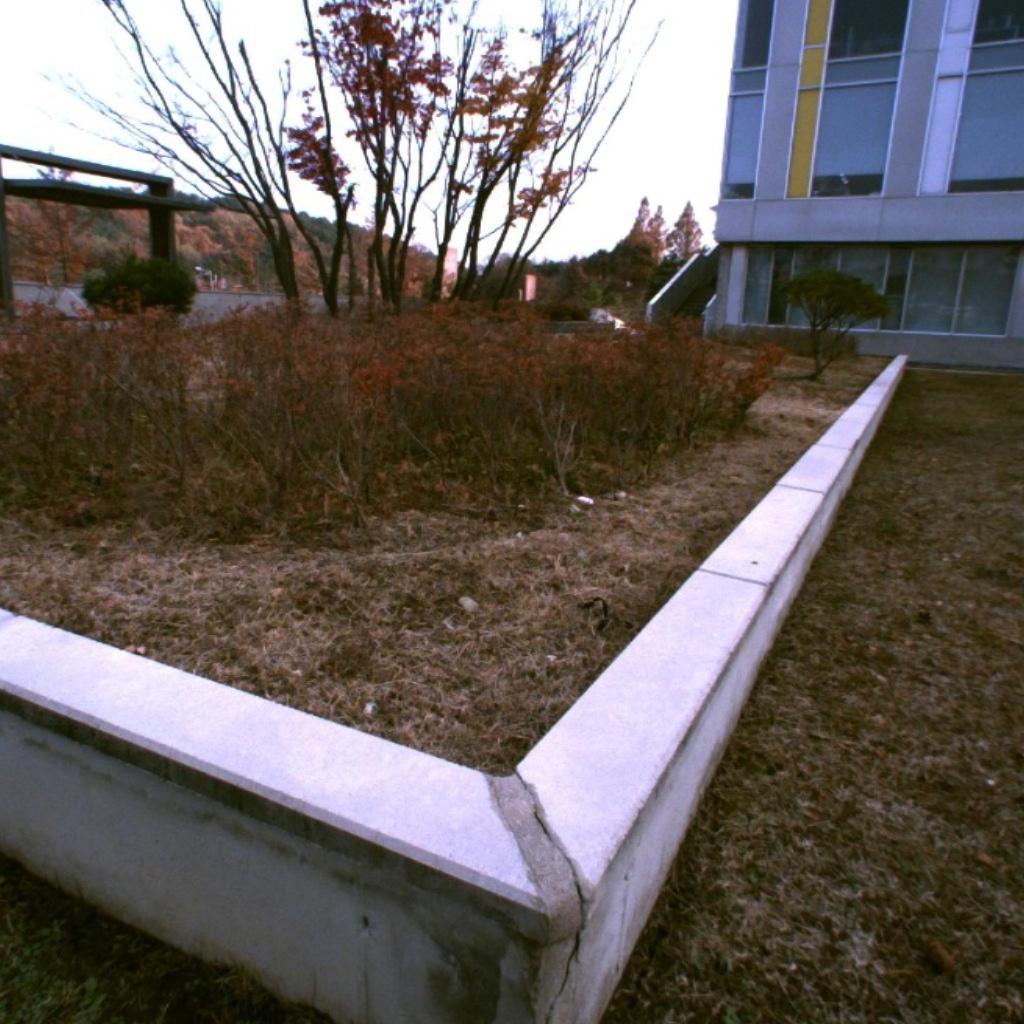} \\
\multicolumn{5}{c}{\footnotesize DRL-AE (Ours)}\\
\includegraphics[width=0.18\linewidth]{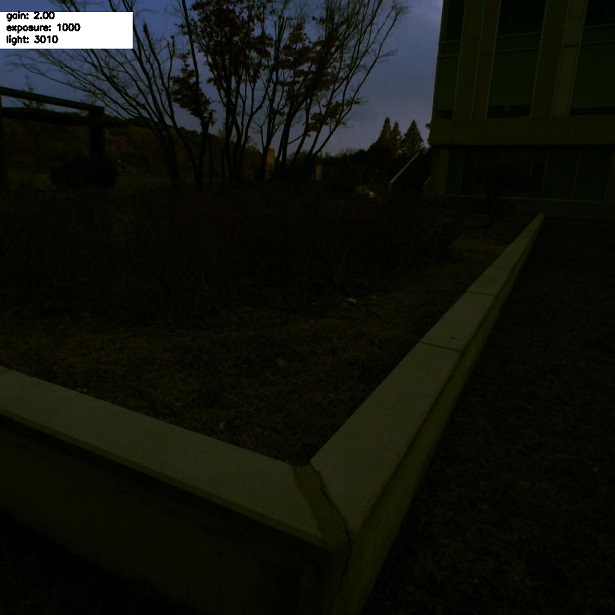} &
\includegraphics[width=0.18\linewidth]{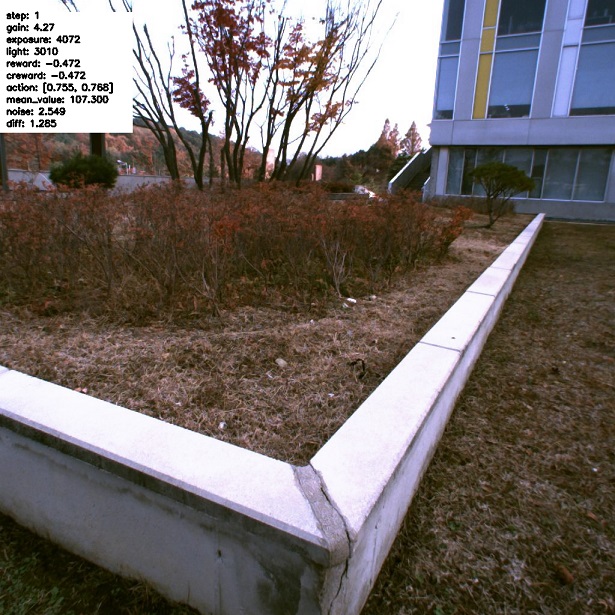} &
\includegraphics[width=0.18\linewidth]{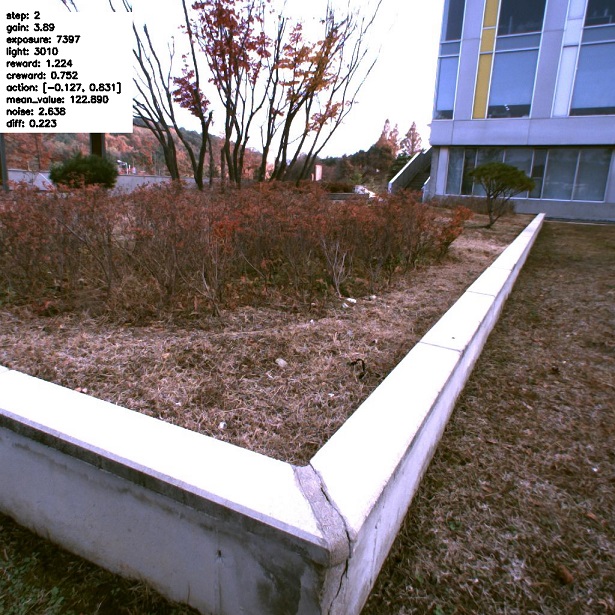} &
\includegraphics[width=0.18\linewidth]{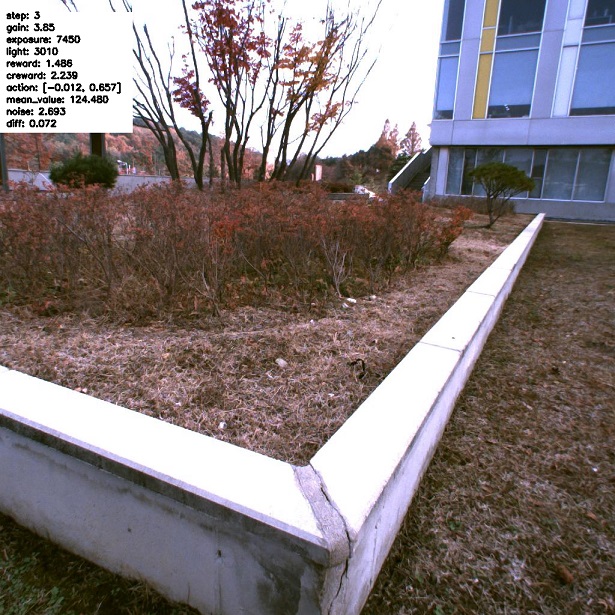} &
\includegraphics[width=0.18\linewidth]{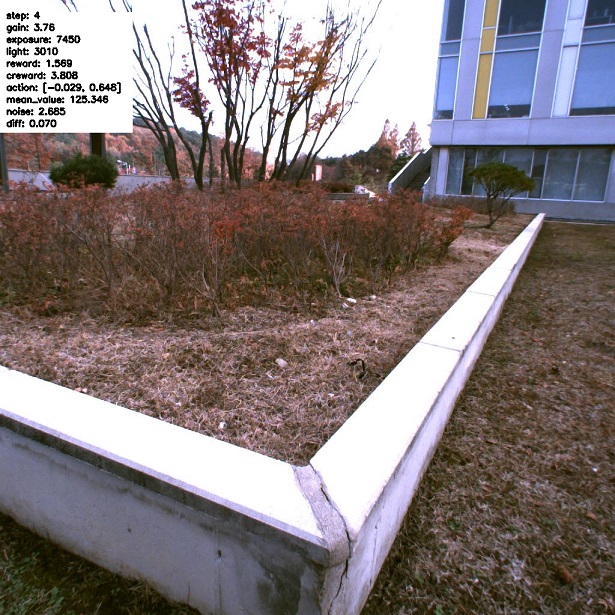} \\
\multicolumn{5}{c}{\footnotesize (a) Acquired image comparison for each convergent step}\\
\end{tabular}
\begin{tabular}{c}
\includegraphics[width=0.90\linewidth]{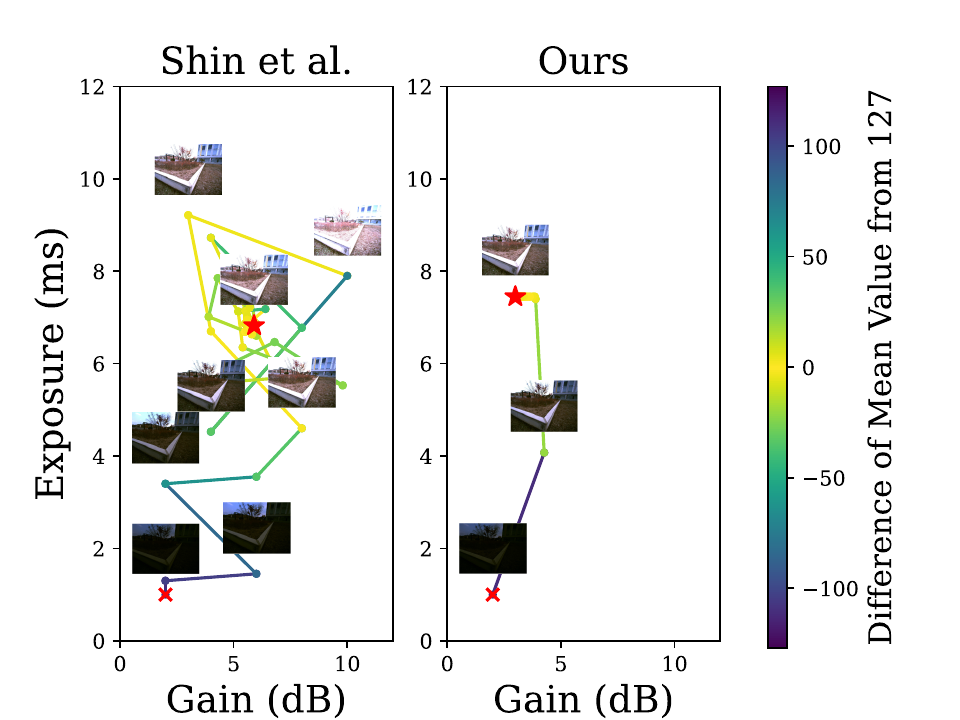} \\
{\footnotesize (b) Convergence trajectory of exposure parameter optimization}
\end{tabular}
\end{center}
\vspace{-0.1in}
\caption{{\bf Convergent step comparison in exposure control dataset~\cite{shin2019camera}.} 
Within three frames, our method already reaches a well-exposed image (a) with minimum exploration (b). 
On the other hand, Shin \etal~\cite{shin2019camera} search local areas with multiple steps (about 30 frames) to converge.
}
\label{fig:simulated}
\end{figure}

\begin{figure}[t]
\begin{center}
\centering
\begin{tabular}{c@{\hskip 0.005\linewidth}c@{\hskip 0.005\linewidth}c@{\hskip 0.005\linewidth}c@{\hskip 0.005\linewidth}c}
\toprule
\multicolumn{5}{c}{\footnotesize Indoor Environment} \\
\hline
\multicolumn{5}{c}{\footnotesize DRL-AE (Ours)} \\
\includegraphics[width=0.18\linewidth]{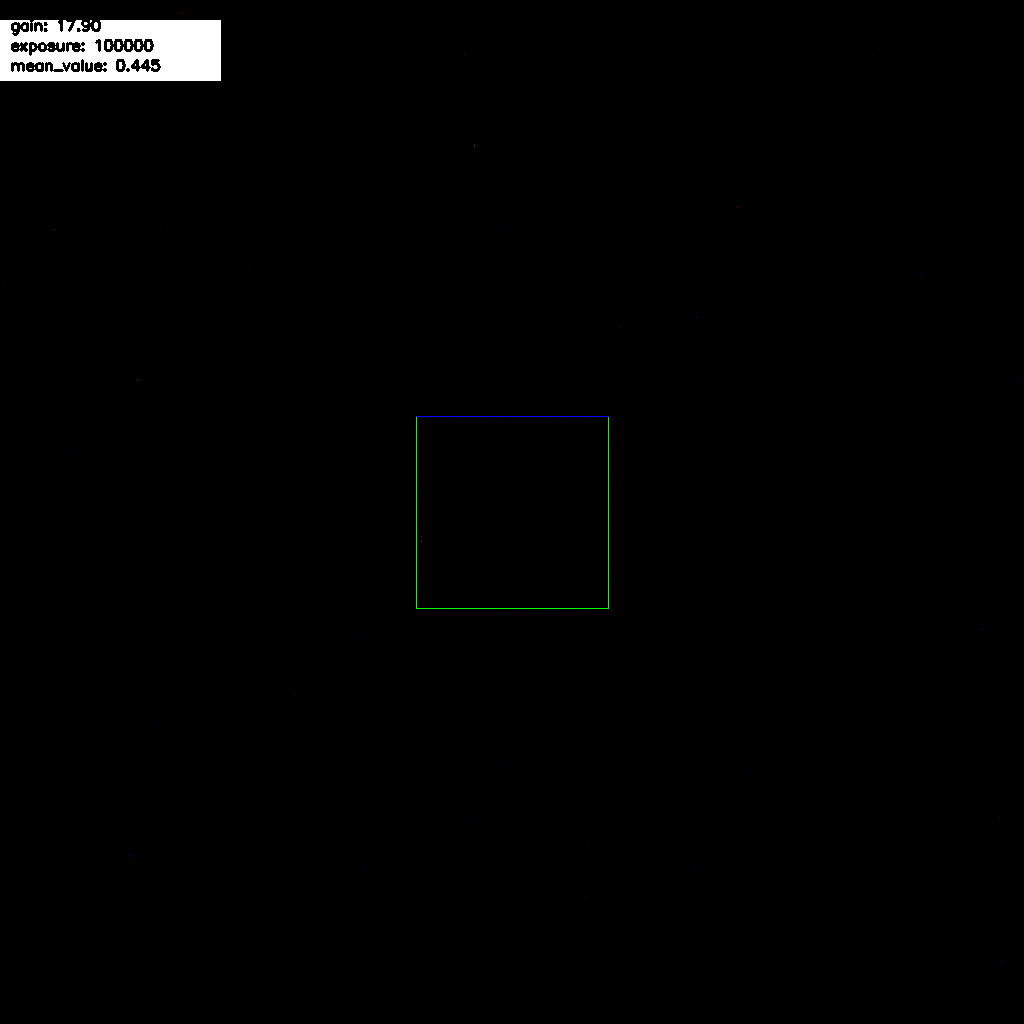} &
\includegraphics[width=0.18\linewidth]{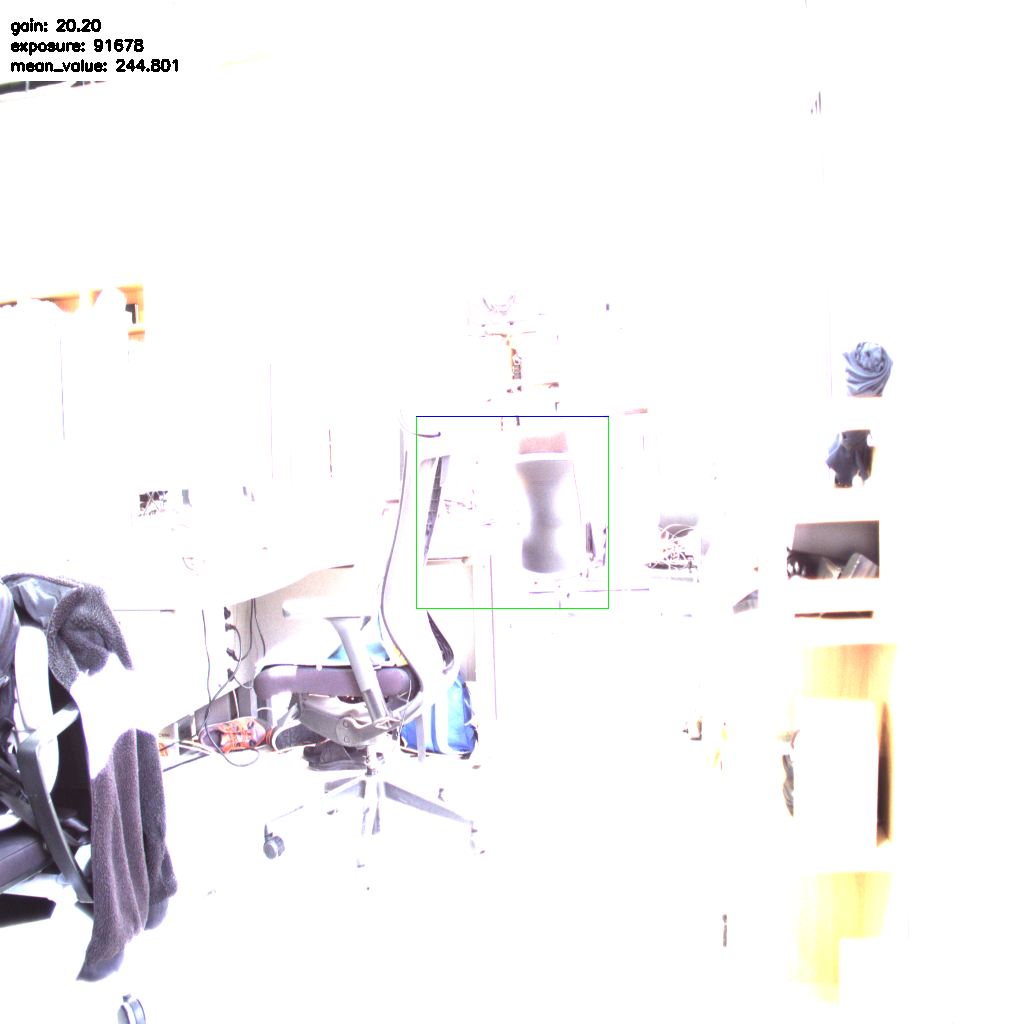} &
\includegraphics[width=0.18\linewidth]{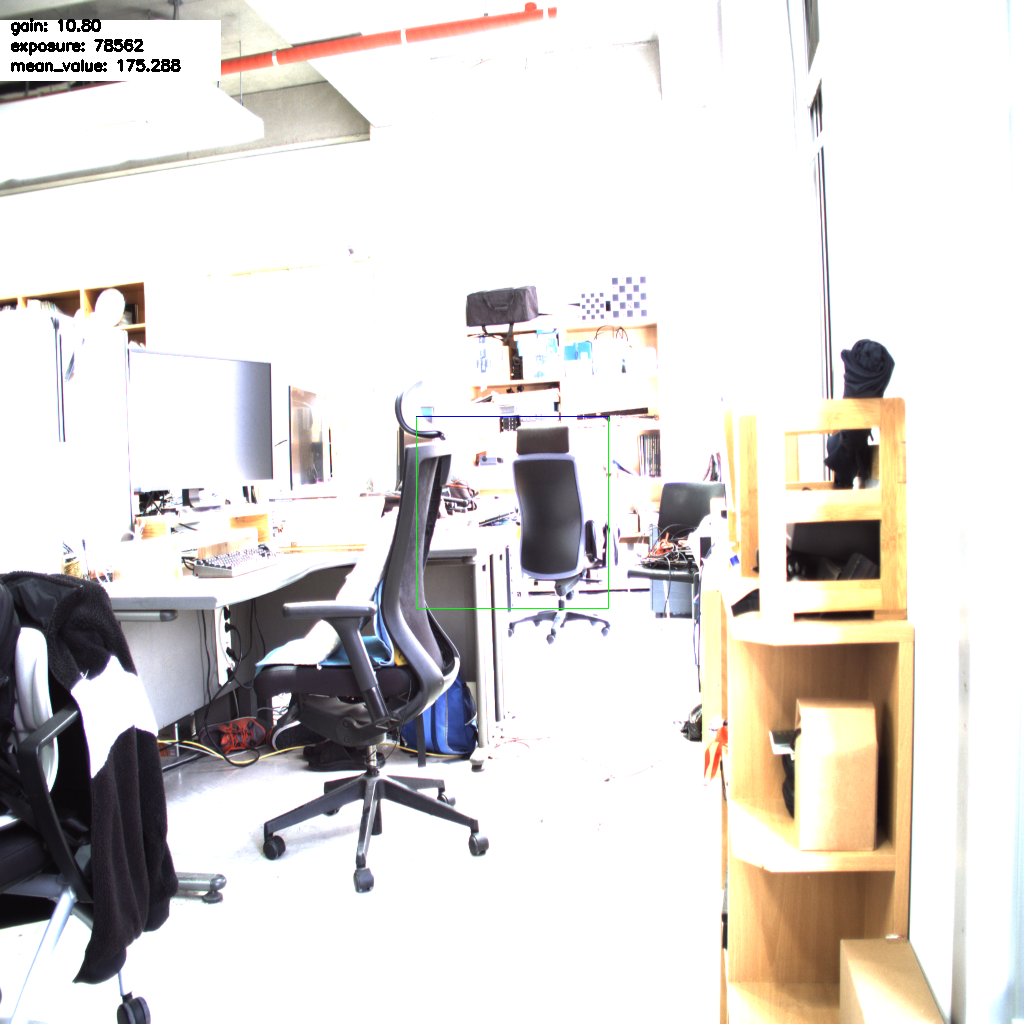} &
\includegraphics[width=0.18\linewidth]{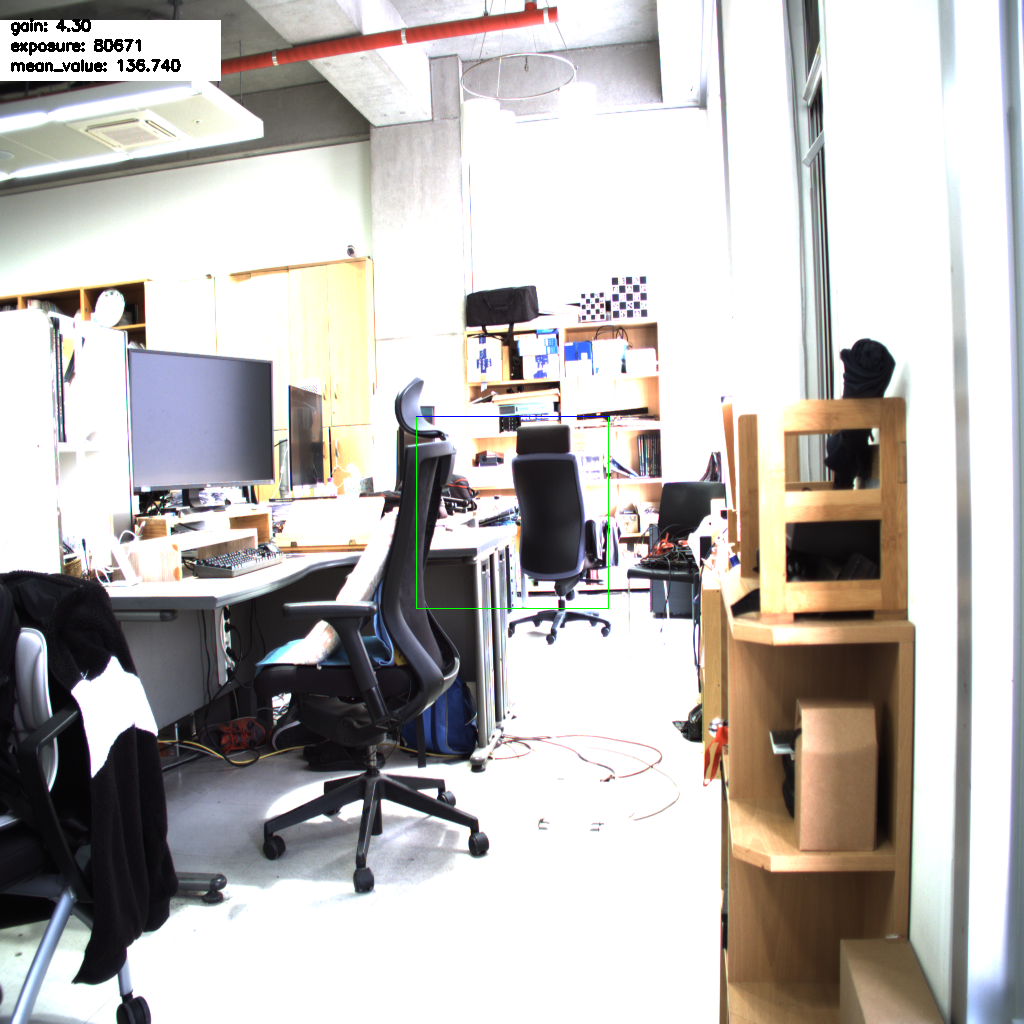} &
\includegraphics[width=0.18\linewidth]{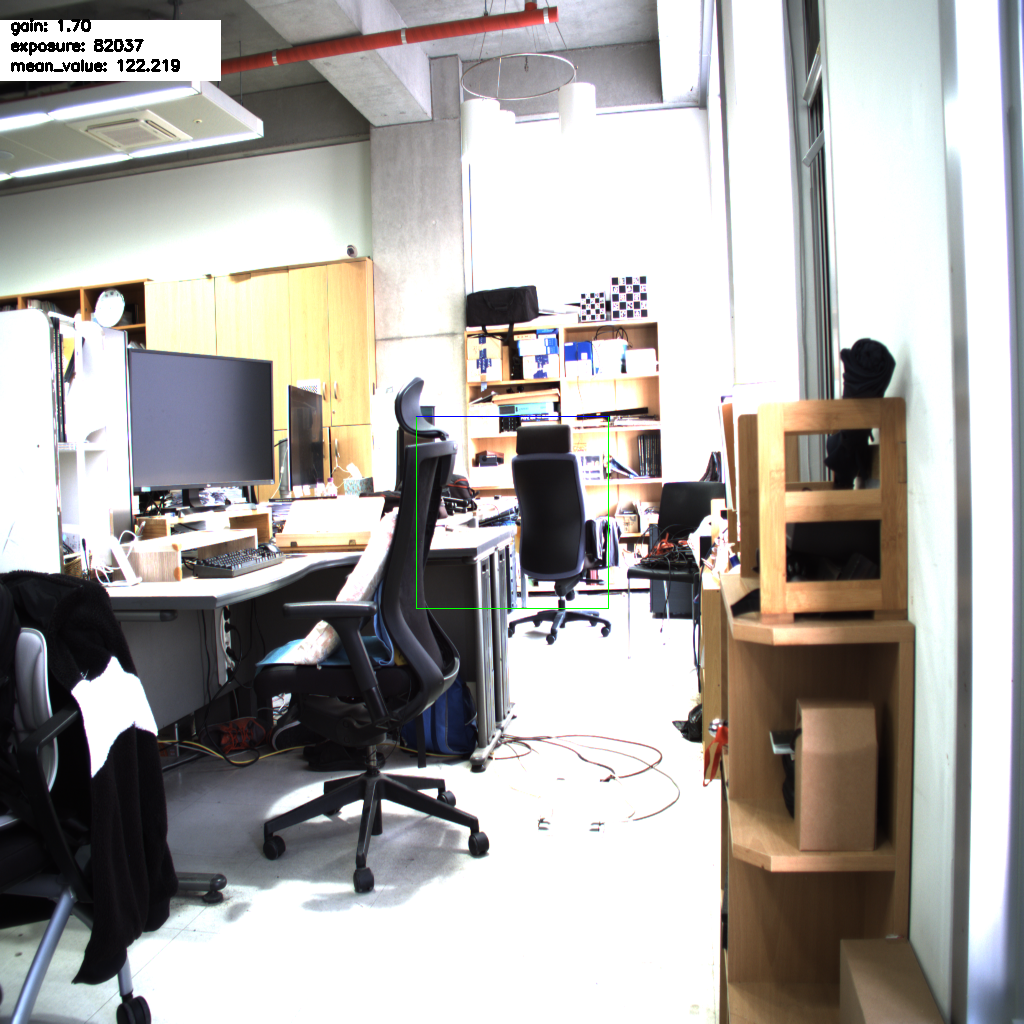} \\
\hline
\multicolumn{5}{c}{\footnotesize Built-in AE} \\
\includegraphics[width=0.18\linewidth]{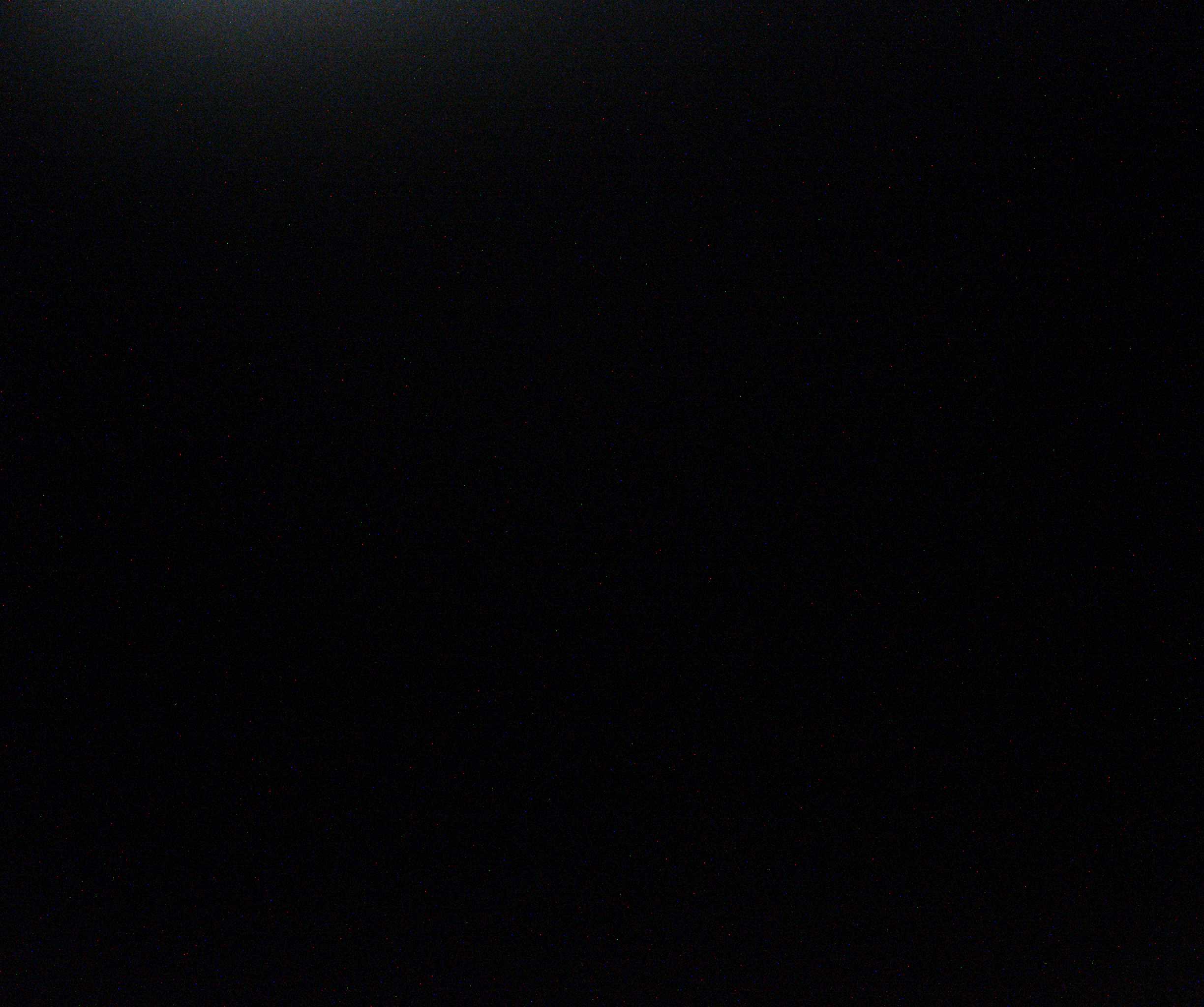} &
\includegraphics[width=0.18\linewidth]{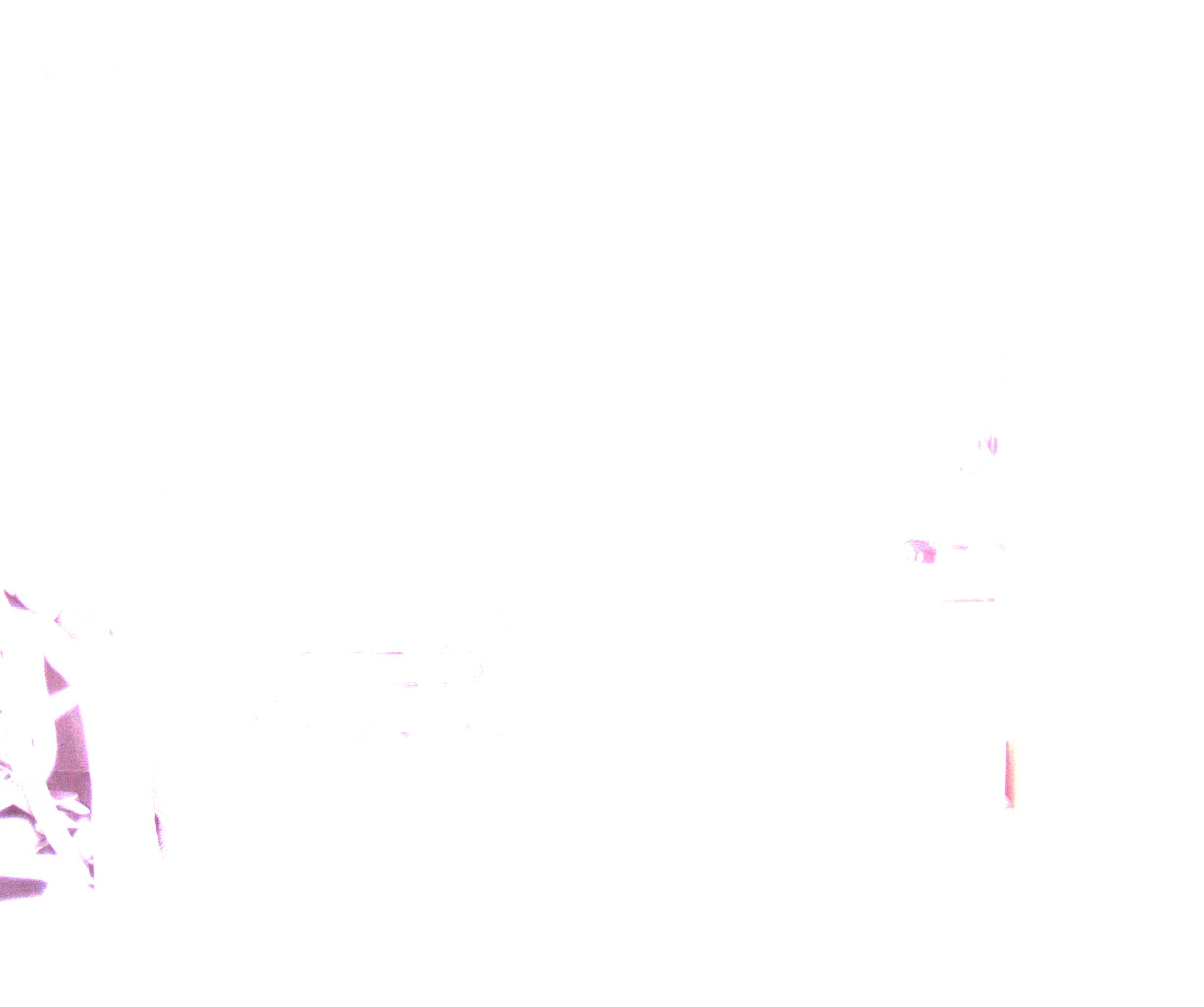} &
\includegraphics[width=0.18\linewidth]{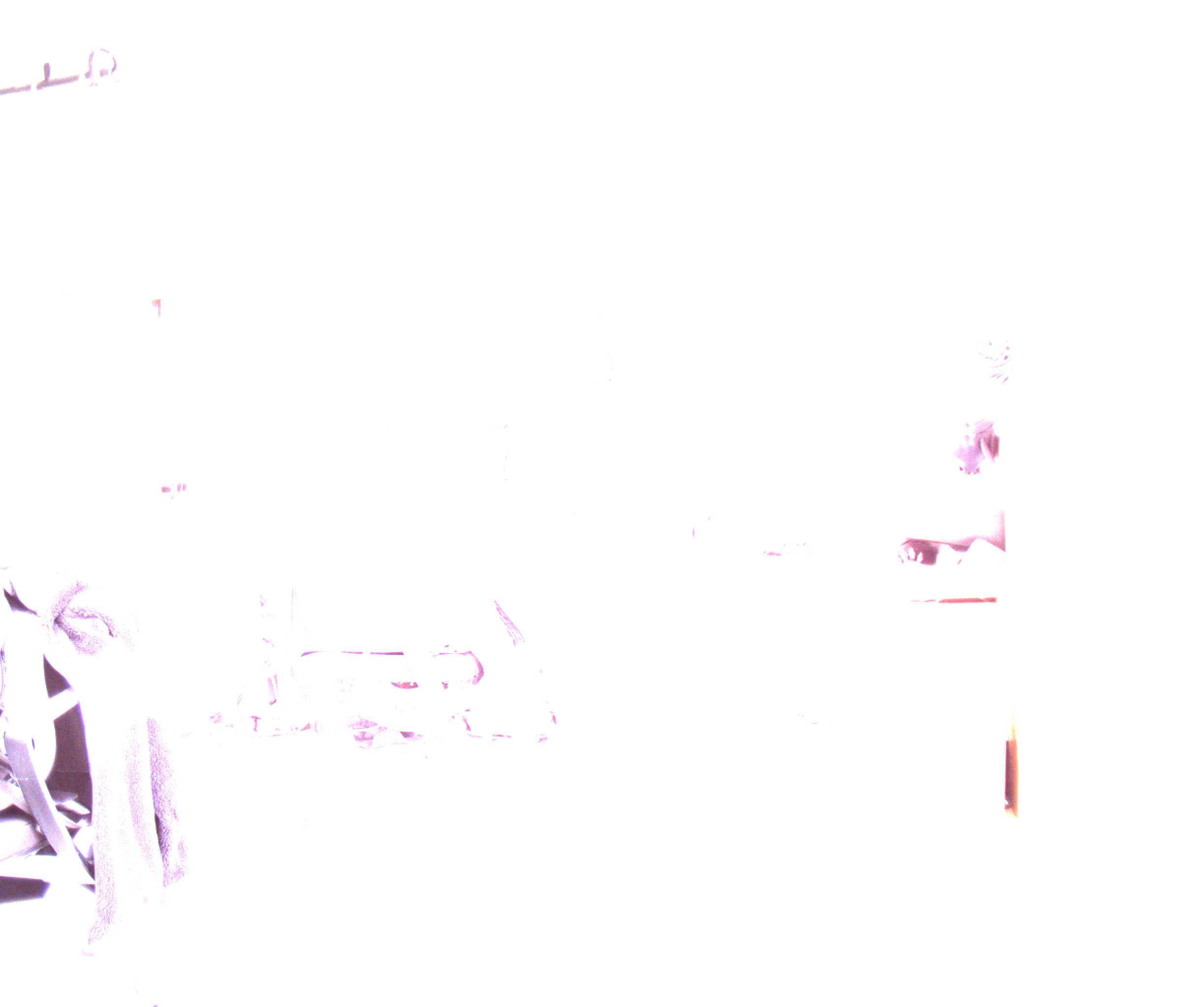} &
\includegraphics[width=0.18\linewidth]{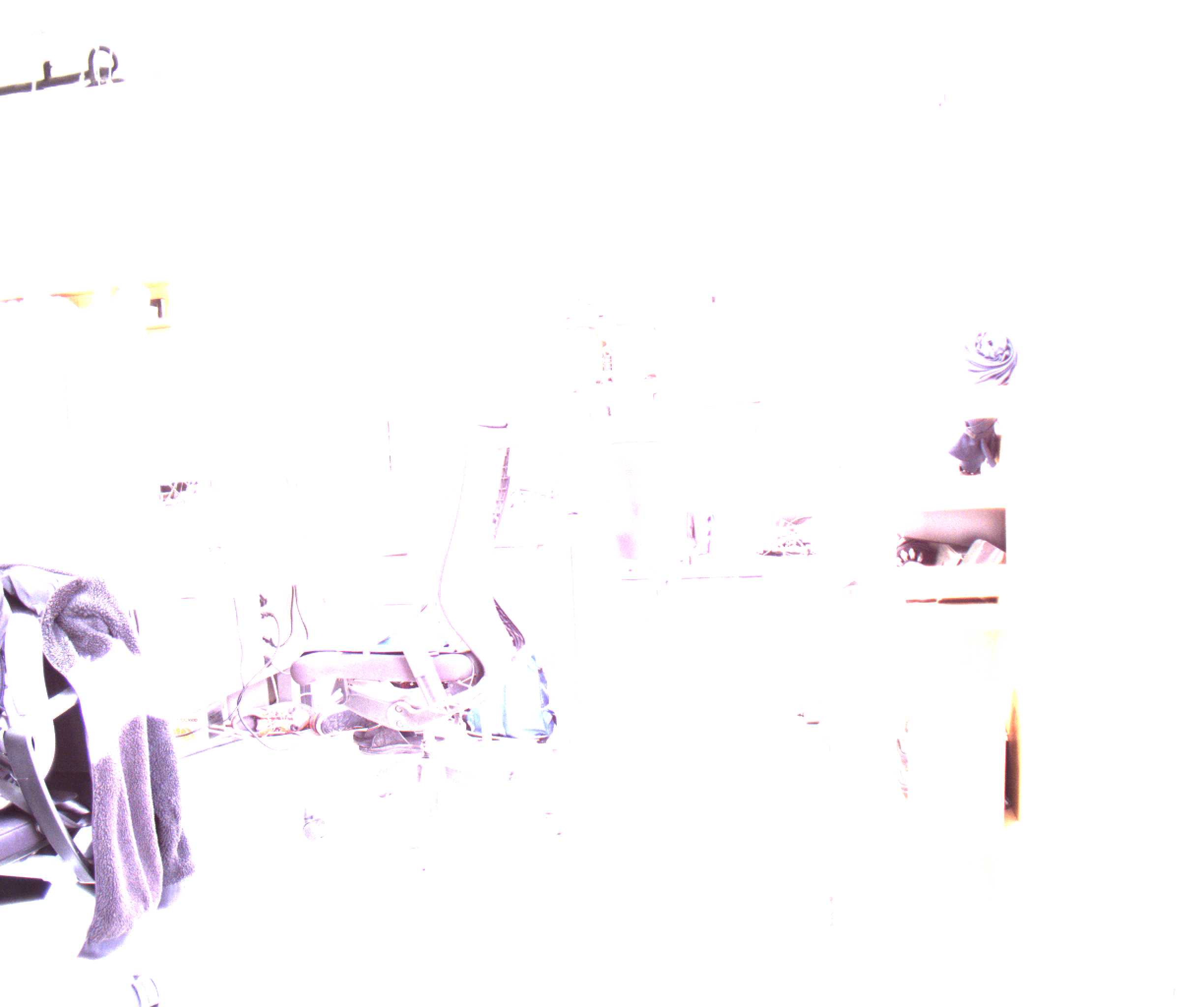} &
\includegraphics[width=0.18\linewidth]{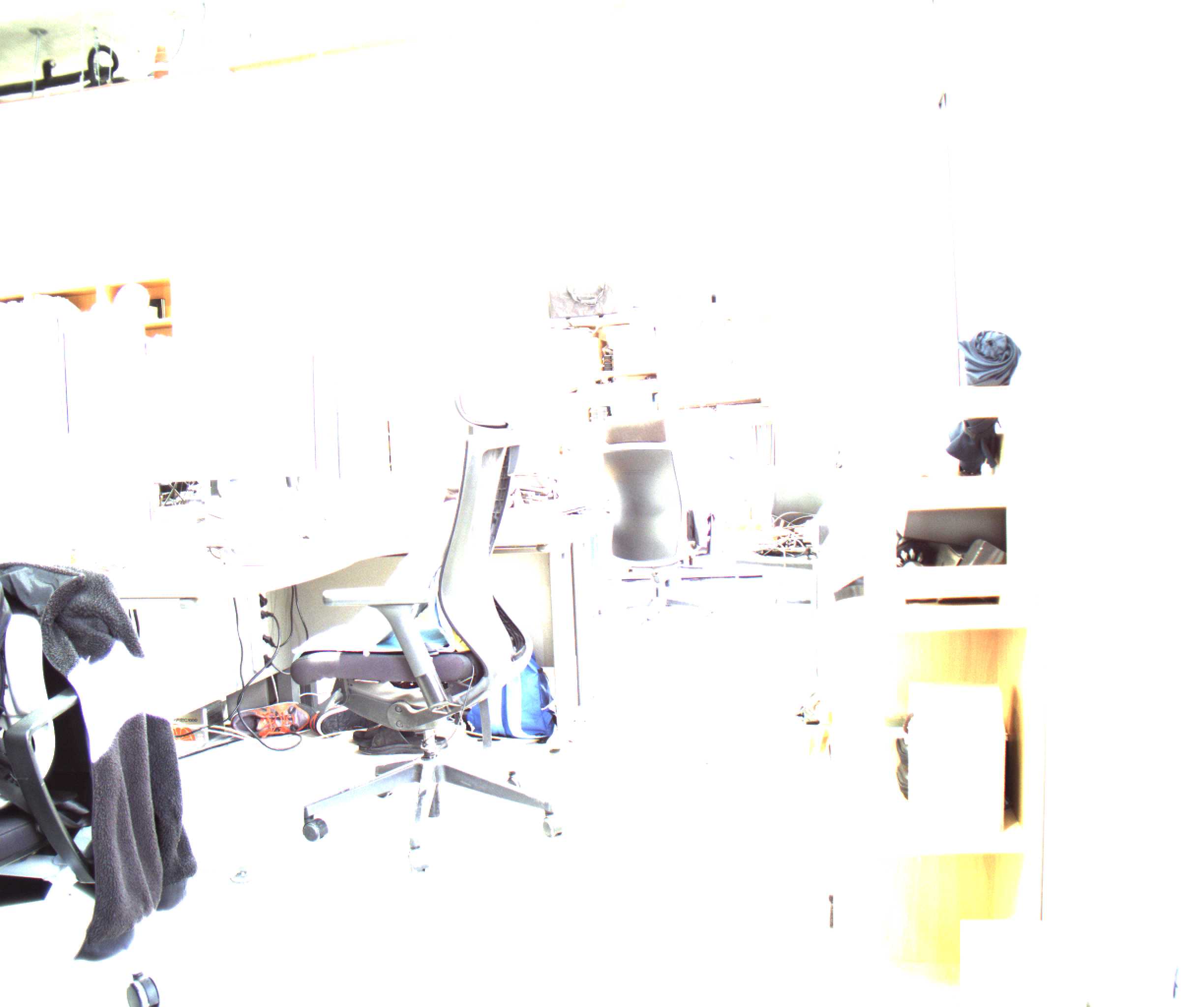} \\
\midrule
\midrule
\multicolumn{5}{c}{\footnotesize Outdoor Environment} \\
\hline
\multicolumn{5}{c}{\footnotesize DRL-AE (Ours)} \\
\includegraphics[width=0.18\linewidth]{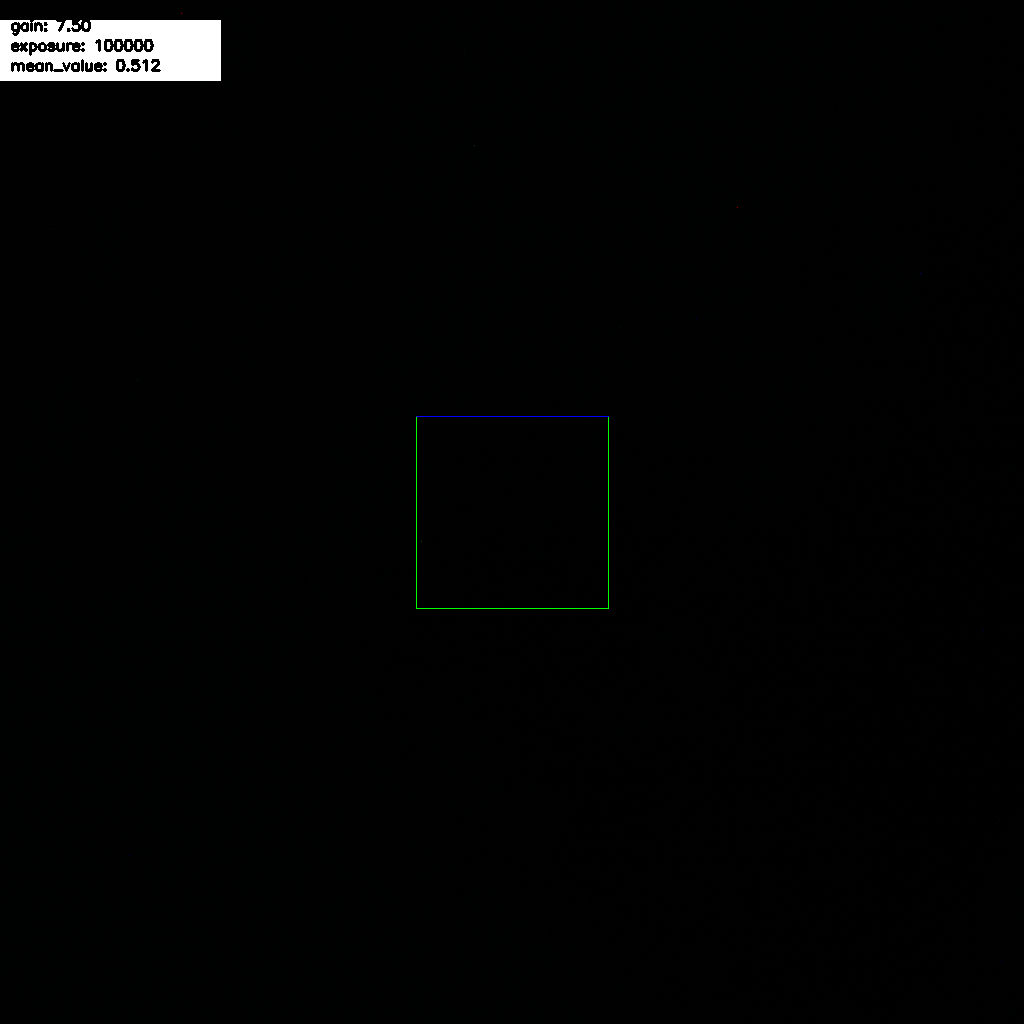} &
\includegraphics[width=0.18\linewidth]{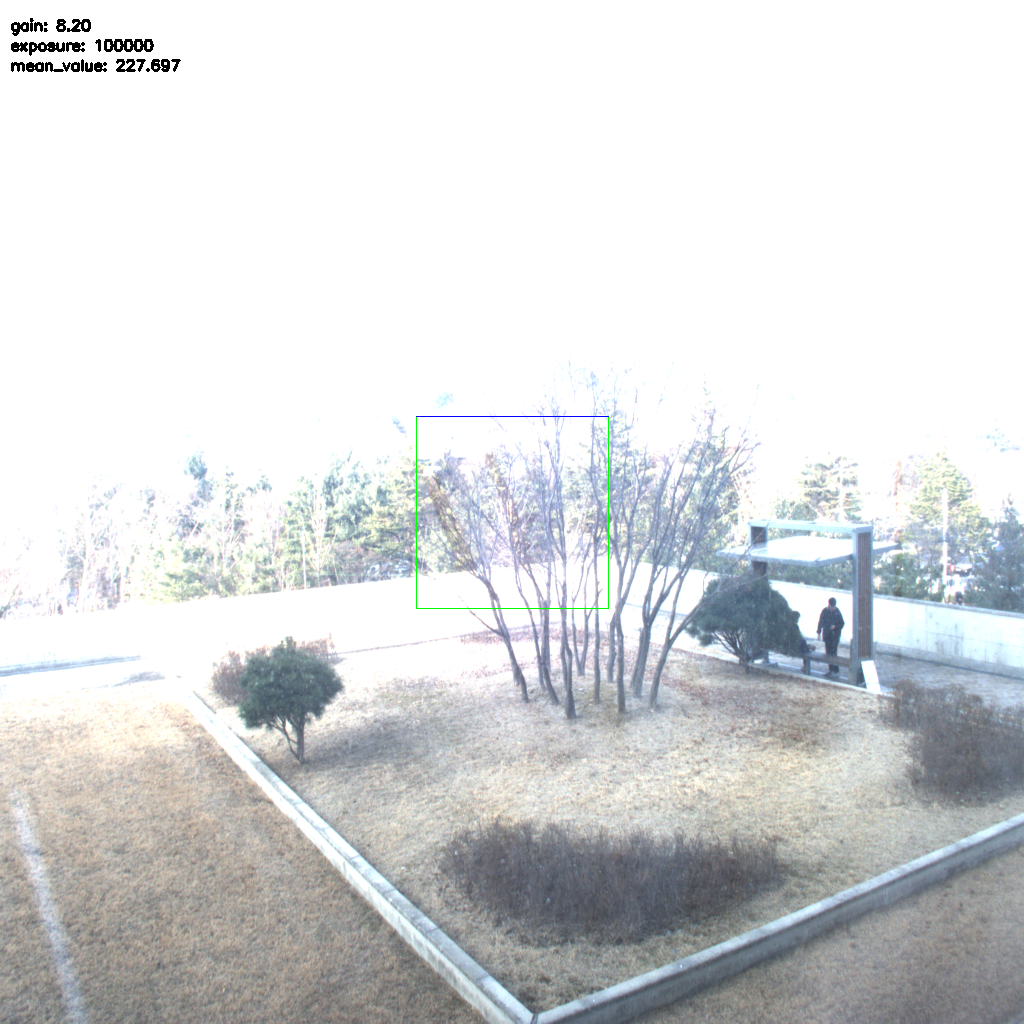} &
\includegraphics[width=0.18\linewidth]{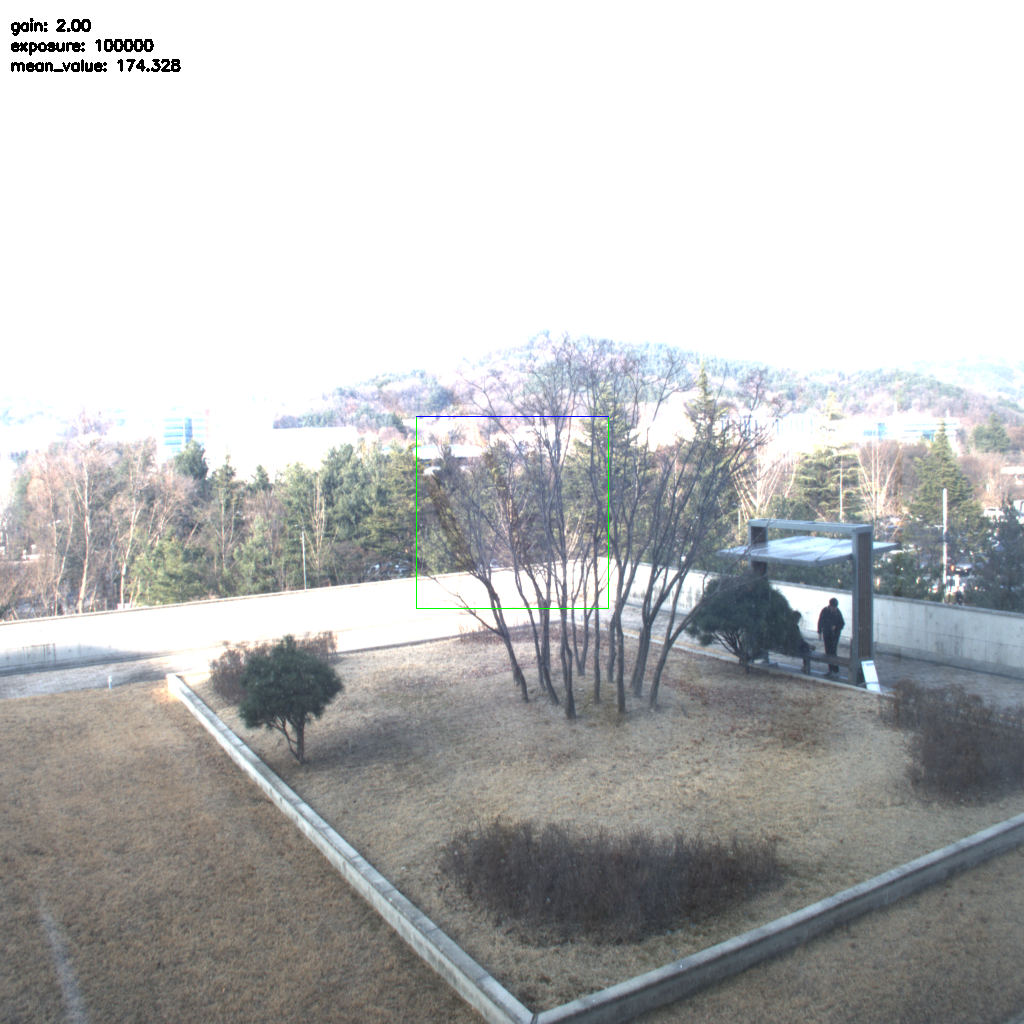} &
\includegraphics[width=0.18\linewidth]{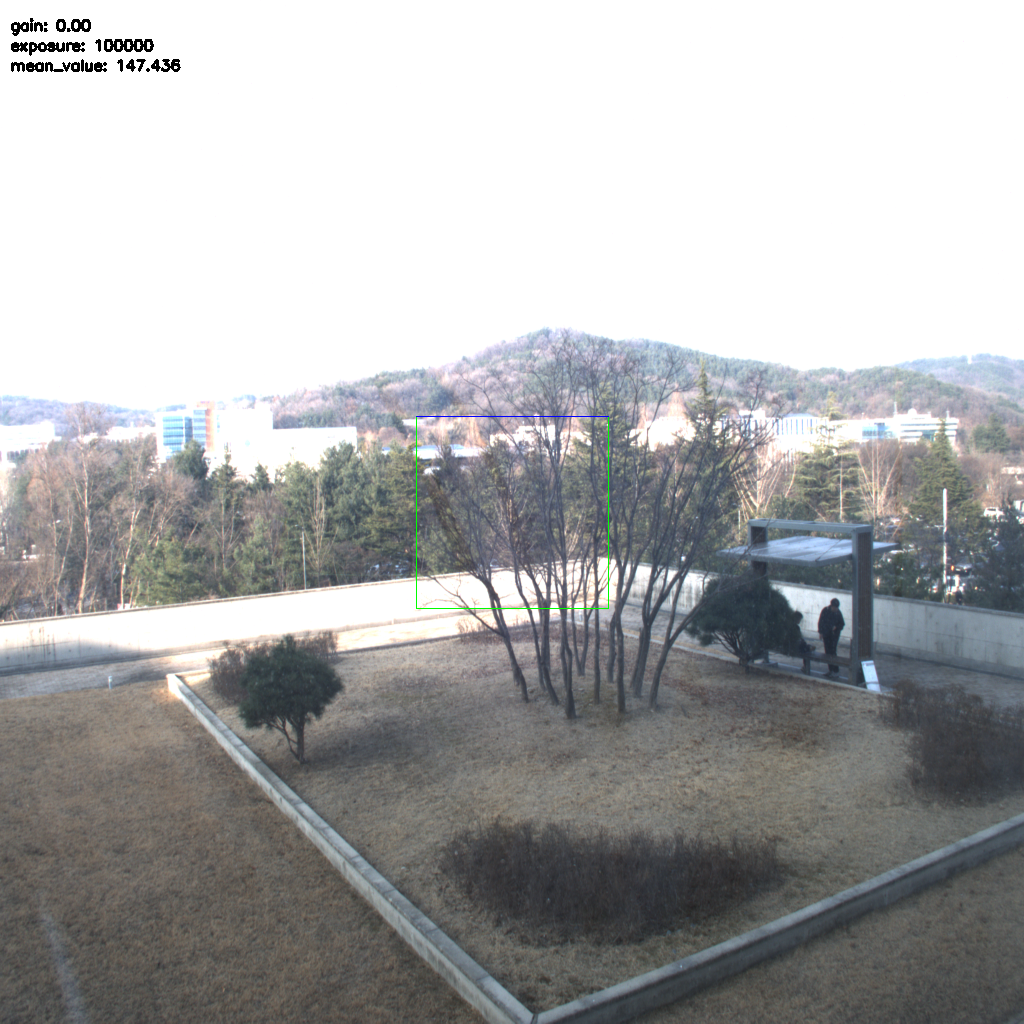} &
\includegraphics[width=0.18\linewidth]{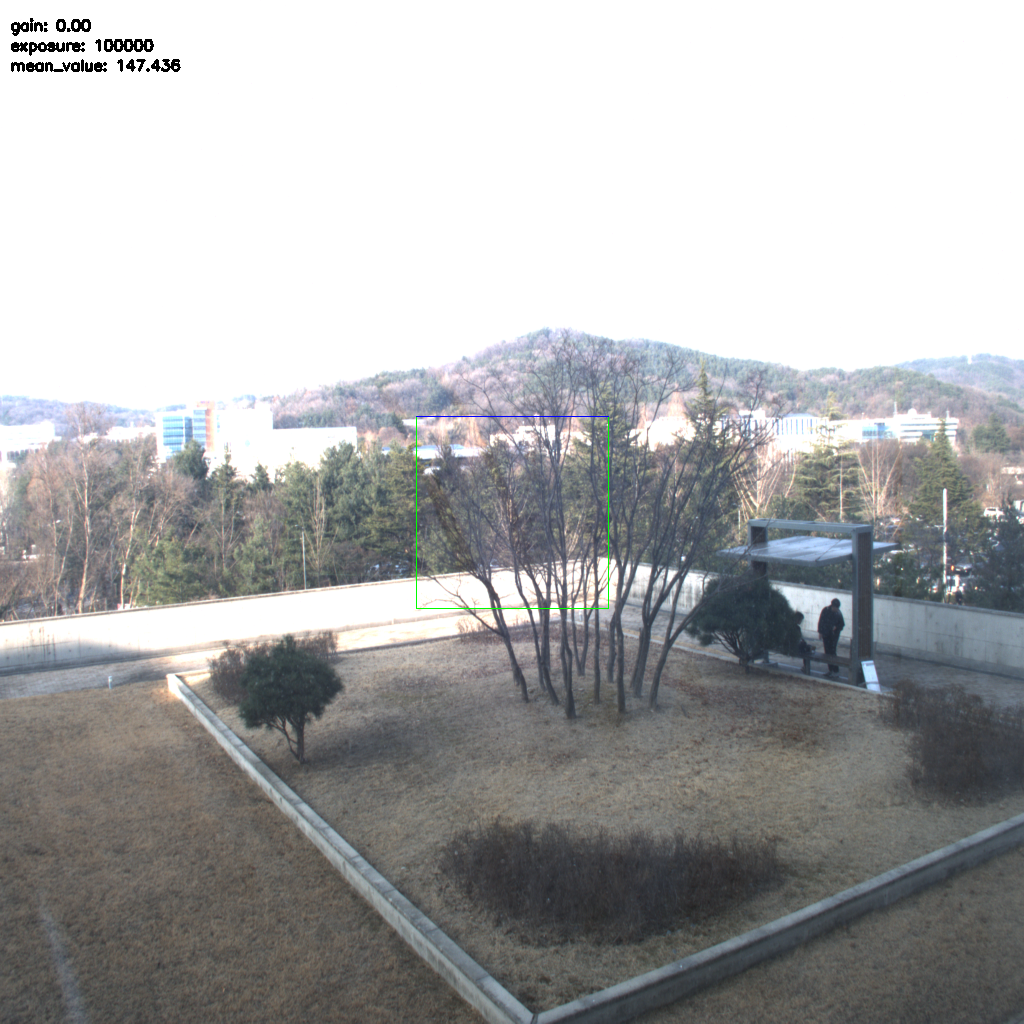} \\
\hline
\multicolumn{5}{c}{\footnotesize Built-in AE} \\
\includegraphics[width=0.18\linewidth]{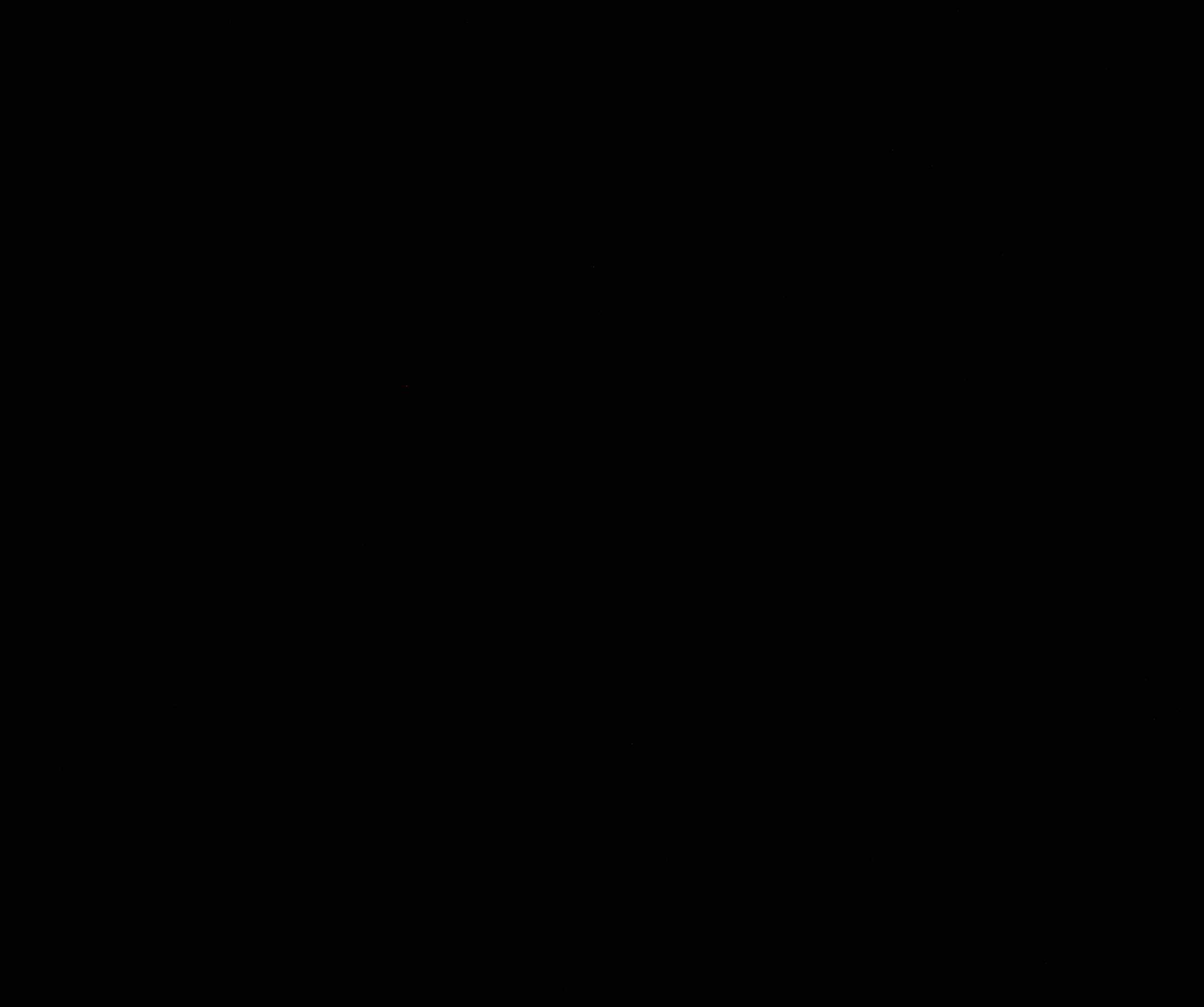} &
\includegraphics[width=0.18\linewidth]{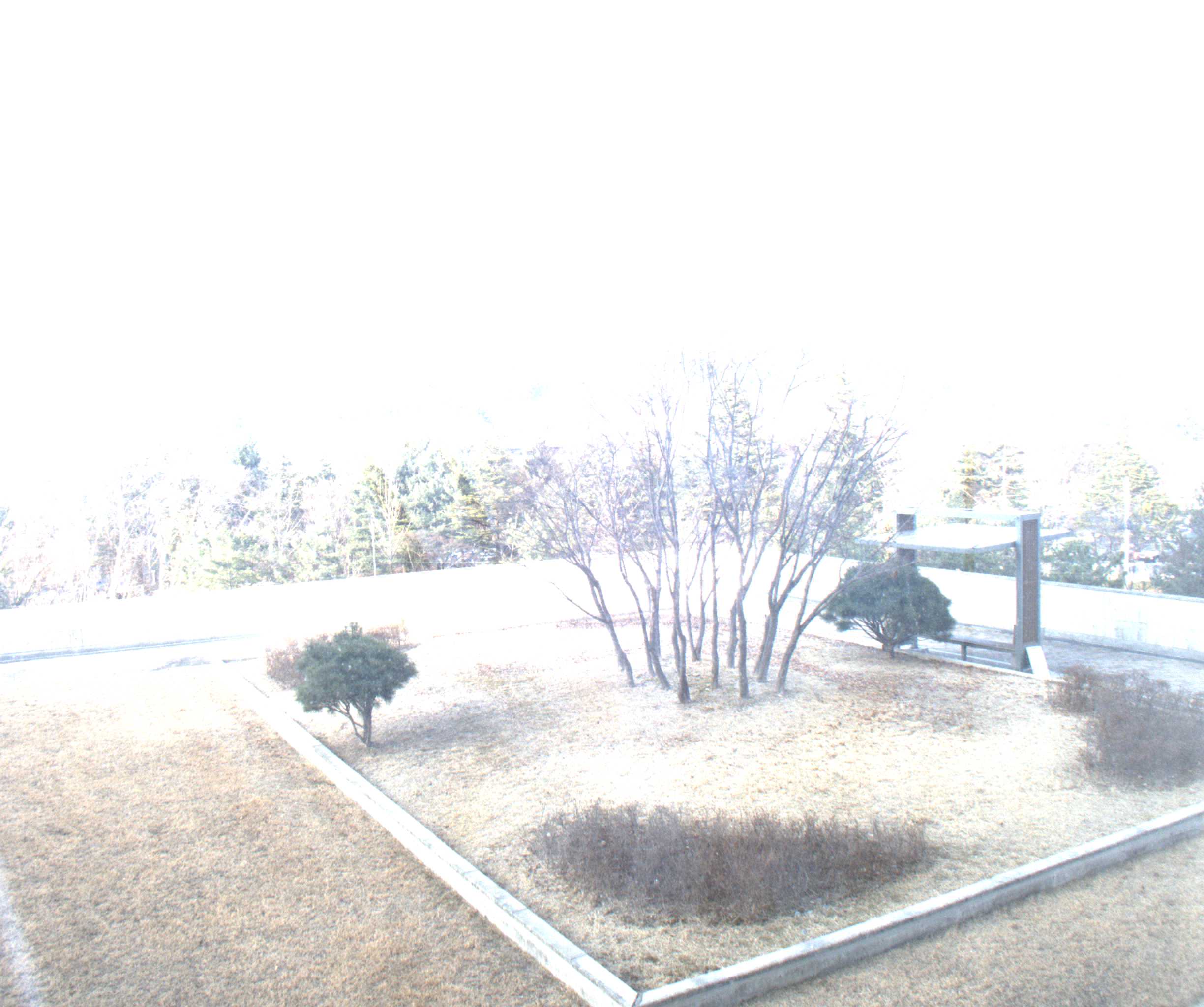} &
\includegraphics[width=0.18\linewidth]{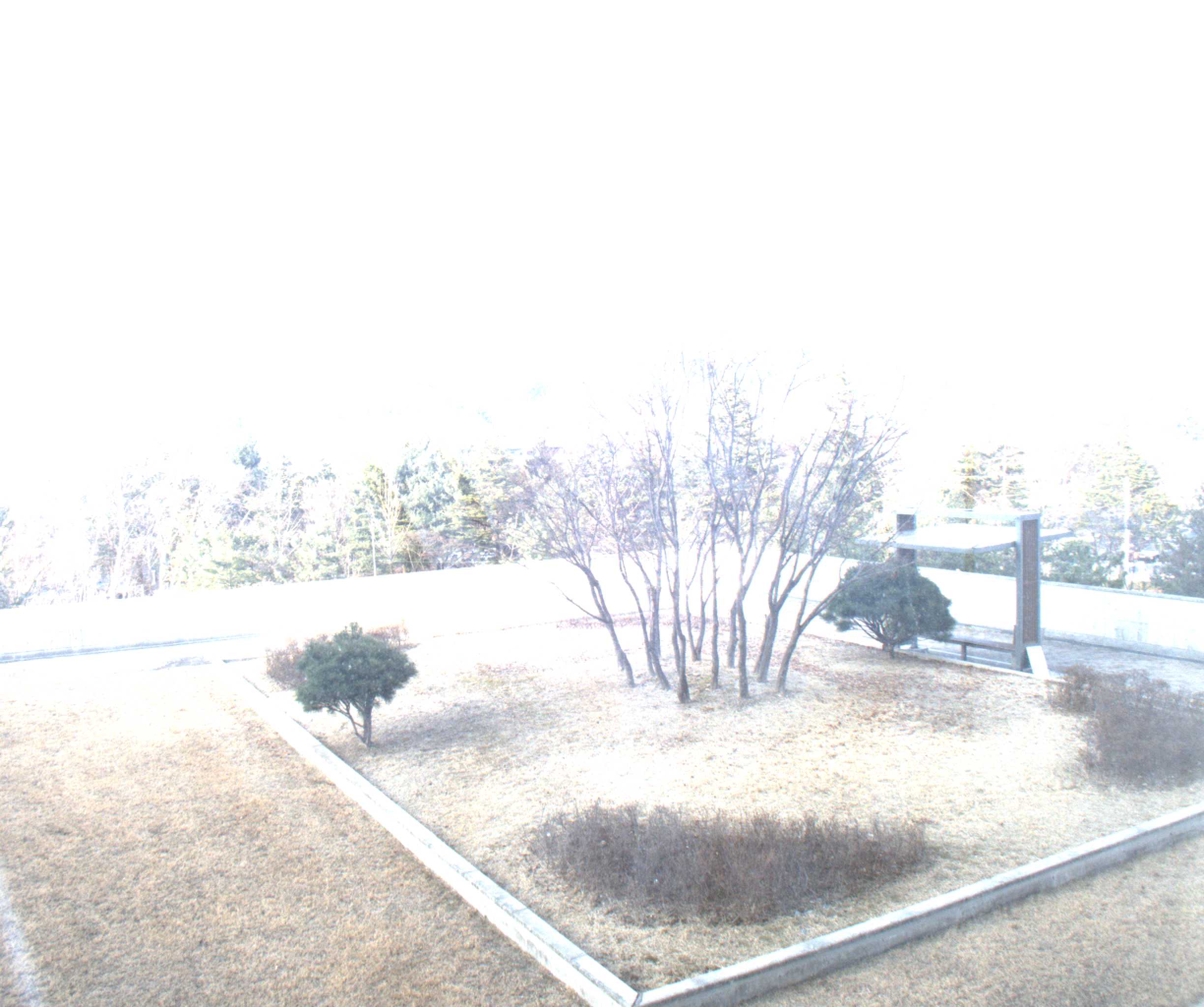} &
\includegraphics[width=0.18\linewidth]{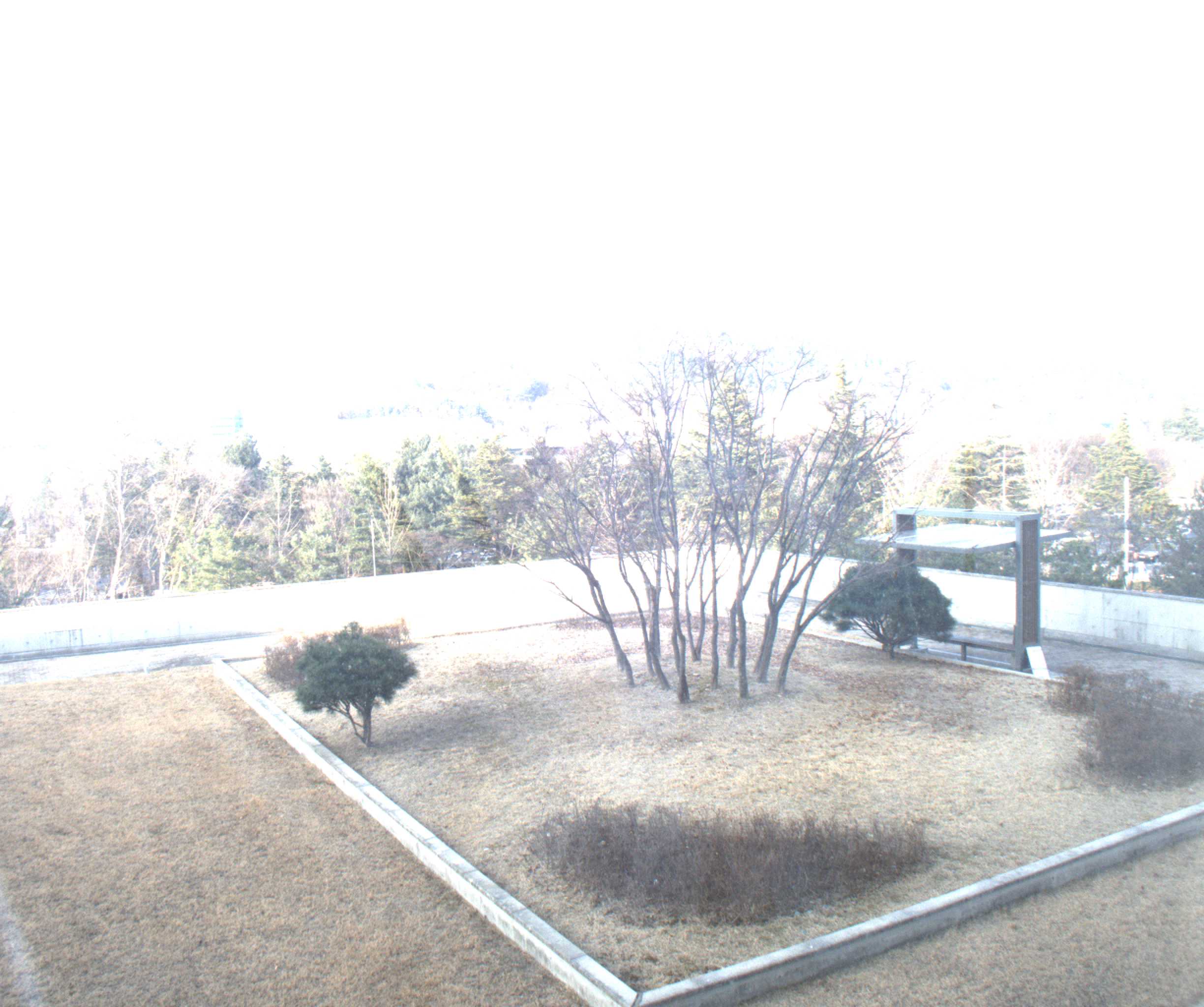} &
\includegraphics[width=0.18\linewidth]{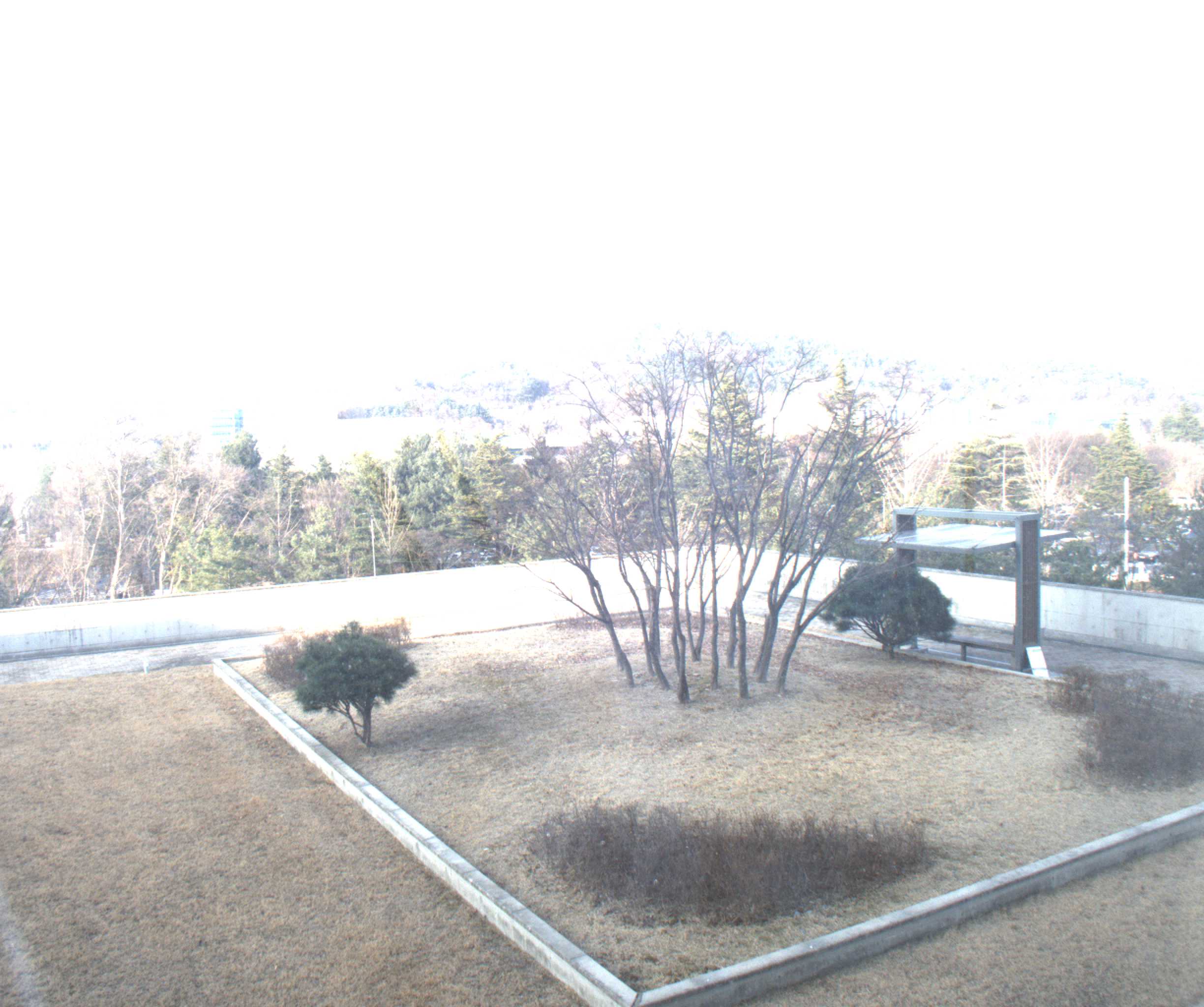} \\
\bottomrule
\end{tabular}
\end{center}
\caption{{\bf Real-world generalization.} 
We compare our method with the camera's built-in exposure control algorithm in real-world scenarios.
Camera lenses are occluded at the initial and suddenly removed in the first frame. 
Our agent converges to a well-exposed image within 3-5 frames. 
Yet, the built-in AE algorithm is still in the middle of adjusting the exposure parameters and is far from the well-exposed image, especially in the indoor case. 
Note that our agent is only trained in the light-controlled darkroom, and this is the zero-shot inference result in the wild.
}
\label{fig:rlvsaeinreal}
\vspace{-0.1in}
\end{figure}

\subsection{Convergent Step Comparison}

\quad \textbf{Exposure Control Dataset~\cite{shin2019camera}.}
The dataset provides multiple images with many different pairs of exposure and gain values, which are captured from the real world.
The dataset consists of several locations, including indoor and outdoor places, with a wide range of exposure and gain values. 
Outdoor images can have the exposure time from $100\mu s$ to $7450 \mu s$, at intervals of $150 \mu s$, and the gain from $0 dB$ to $20 dB$ with $2 dB$ interval. 
Similarly, indoor images have exposure time from $4 ms$ to $67 ms$ with $3 ms$ intervals and gain from $0 dB$ to $24 dB$ with $2 dB$ intervals.

We evaluate our method with \citet{shin2019camera}.
The final converged points are slightly different because of the difference between the proposed reward function and the assessment metric of \cite{shin2019camera}. 
Both algorithms converge to a comparable point, as shown in \figref{fig:simulated}.
It only takes three frames to converge with our method. 
However, the Nelder-Mead optimization method in \cite{shin2019camera} takes at least 30 timesteps to converge completely.
Therefore, it is hard to use it in real scenarios, although they may find a more optimal point. 

\textbf{Real-world Indoor and Outdoor Environment.}
We evaluate our method with the camera's built-in AE control algorithm in real-world scenarios. 
The purposes of this experiment are twofold: 1) comparing convergence speed with the built-in AE algorithm, and 2) testing zero-shot generalization performance in the wild.

Before starting, we cover the camera lens with enough time to converge in the dark, then quickly remove it to test the convergence speed for a sudden lighting change.
\figref{fig:rlvsaeinreal} shows the captured initial five images during each optimization. 
Our method converges to a well-exposed image within 3-5 frames in both indoor and outdoor scenes.
However, the built-in AE algorithm takes much longer to converge: 30 frames for indoors and 10 frames for outdoors. 

Also, we found that our agent shows satisfactory zero-shot generalization performance, even though it is only trained in the light-controlled darkroom with limited object context. 
We believe our state design (\ie, vectorized intensity history) and spatial domain randomization bring this result by removing the potential domain gap issue of the CNN feature and augmenting object context as much as possible.

\begin{figure}[t]
\begin{center}
\begin{tabular}{c@{\hskip 0.005\linewidth}c}
\includegraphics[width=0.47\linewidth]{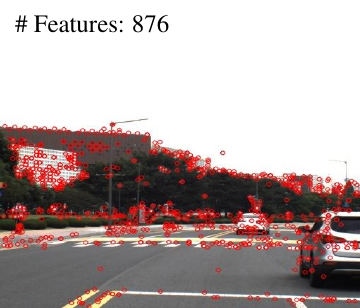} & 
\includegraphics[width=0.47\linewidth]{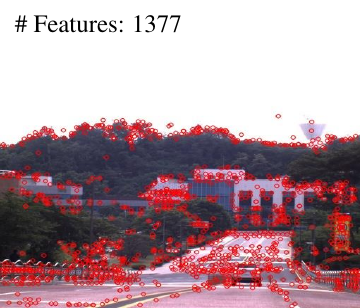} \\
\multicolumn{2}{c}{\footnotesize (a) Feature extraction from DRL-AE (Ours)} \\
\includegraphics[width=0.47\linewidth]{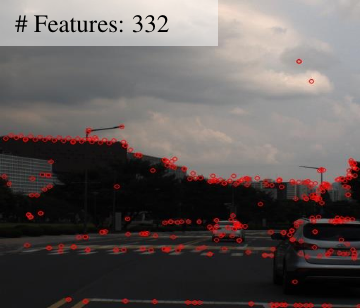} &
\includegraphics[width=0.47\linewidth]{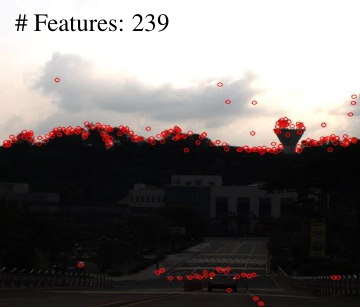} \\
\multicolumn{2}{c}{\footnotesize (b) Feature extraction from built-in AE} \\
\end{tabular}
\end{center}
\vspace{-0.1in}
\caption{{\bf SIFT \cite{lowe1999object} feature extraction result.} Captured images from the proposed algorithm and built-in AE are processed to detect SIFT features. The images were simultaneously captured in real-time from two separate cameras equipped on a driving vehicle. Our method can provide plenty of SIFT features over the image plane. On average, our method detects 38\% more features across a total of 5355 images.}
\label{fig:feature}
\end{figure}

\subsection{Real-time Driving Env: Feature Extraction}
\quad In this experiment, two cameras are attached to the top of a moving car, and images are captured simultaneously. 
One camera is used for our algorithm, and the other camera is for the built-in AE algorithm. 
We tested the algorithms on real-world driving scenarios, including campus and urban roads.
Our algorithm runs on a laptop equipped with an i7-7700HQ@2.80GHz CPU unit.
Given an image, the agent predicts exposure time and gain commands in real-time.
The estimated actions are transmitted to the attached camera.
After driving sequences acquisition, we extract SIFT \cite{lowe1999object} features from each captured image. 
Please note that the images of DRL-AE and built-in AE have slightly different views due to the difference in the installed position. 

\figref{fig:feature} shows the comparison of feature extraction results.
From the total of 5355 image pairs, our method produces 1,306 SIFT features on average with a 1,157 median value.
On the other hand, the built-in AE method only results in 946 features on average, with a 711 median value. 
Therefore, 38\% more features are detected on average, and the difference is up to 62\% for the median value. 
The number of detected features and feature repeatability during exposure transition are critical for Visual Odometry (VO) and SLAM tasks.
So, we believe our method can be valuable for VO, SLAM, and visual tracking tasks as well.

\begin{figure}[t]
\begin{center}
\begin{tabular}{c@{\hskip 0.005\linewidth}c}
\includegraphics[width=0.47\linewidth]{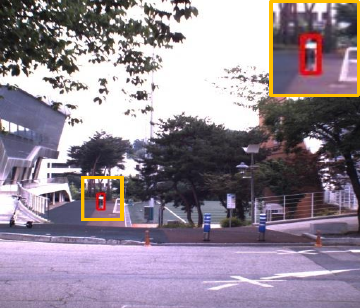} &
\includegraphics[width=0.47\linewidth]{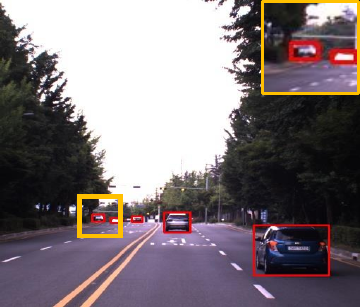} \\
\multicolumn{2}{c}{\footnotesize (a) Object detection from DRL-AE (Ours)} \\
\includegraphics[width=0.47\linewidth]{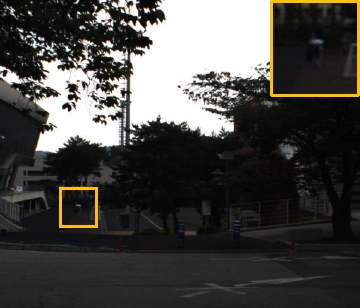} &
\includegraphics[width=0.47\linewidth]{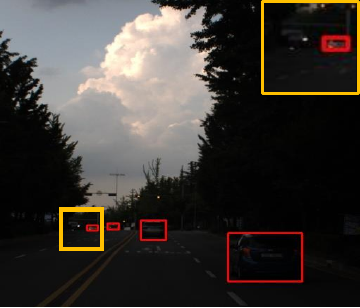} \\
\multicolumn{2}{c}{\footnotesize (b) Object detection from built-in AE} \\
\end{tabular}
\end{center}
\vspace{-0.1in}
\caption{{\bf Object detection result.} 
Captured images from the proposed algorithm and built-in AE are processed to detect target objects.
The experiment used the same image sequence as the SIFT experiment.
We utilize Yolo-v5 \cite{glenn_jocher_2022_7347926} for car and pedestrian detection. 
On average, our method detects 5\% more objects compared to the built-in AE algorithm.
}
\label{fig:detection}
\end{figure}

\subsection{Real-time Driving Env: Object Detection}
\quad Similar to the feature extraction experiment, the captured images are processed with YOLO-v5 \cite{glenn_jocher_2022_7347926}. 
The images are taken from campus and urban road scenes, so we only take into account cars and pedestrians.
\figref{fig:detection} shows the comparison of object detection results from DRL-AE and built-in AE methods. 
Recent detection models, including YOLO-v5, adopt modern augmentation methods to make the model robust the image brightness changes.
Therefore, YOLO-v5 tends to detect objects well in even poorly exposed images.
However, our algorithm detects 5\% more objects in terms of the total number of detected objects. 
Furthermore, the objects in our image are detected much earlier than the built-in AE method. 
This is highly critical for autonomous vehicles that are driving at high speed.
Earlier detected objects can prevent human injury and potential accidents.

\begin{figure}[t]
\begin{center}
\begin{tabular}{c@{\hskip 0.005\linewidth}c@{\hskip 0.005\linewidth}c}
\includegraphics[width=0.32\linewidth]{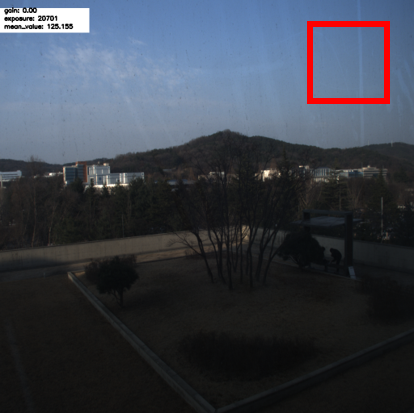} &
\includegraphics[width=0.32\linewidth]{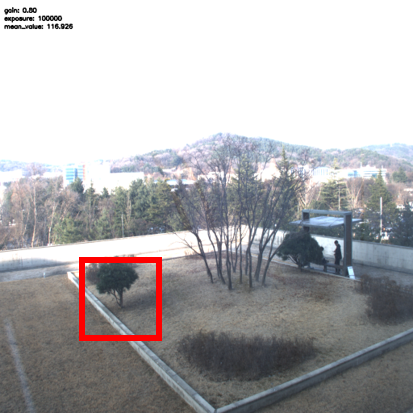} &
\includegraphics[width=0.32\linewidth]{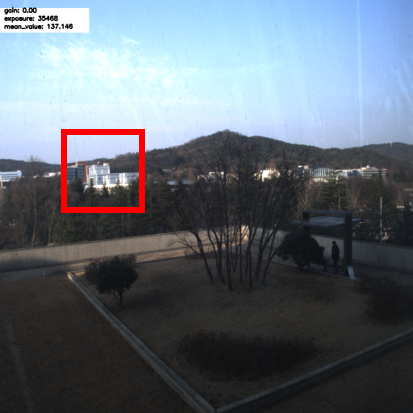} \\
\includegraphics[width=0.32\linewidth]{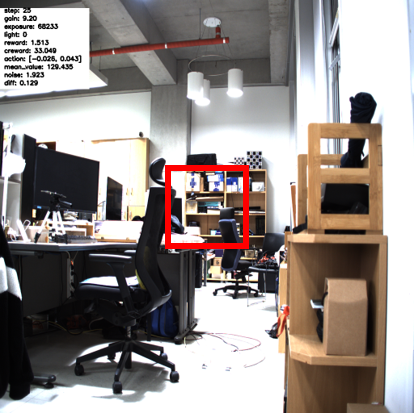} &
\includegraphics[width=0.32\linewidth]{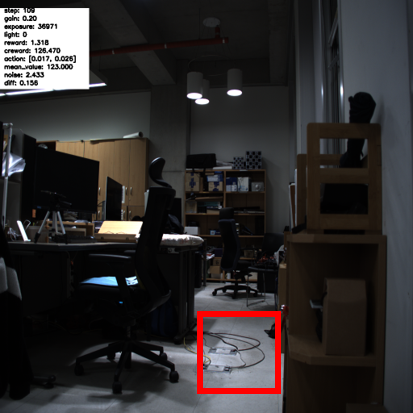} & 
\includegraphics[width=0.32\linewidth]{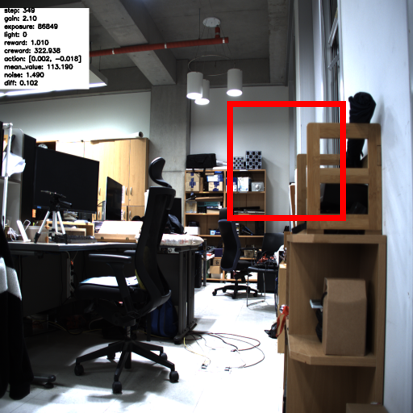} \\
\end{tabular}
\end{center}
\caption{{\bf RoI-aware camera exposure control.} Our agent is able to control camera exposure for a specific RoI or entire image. Given a RoI box, the agent adjusts the camera parameters to maximize the image attribute for a specific RoI area. It allows the camera to capture the detailed context for the regions of interest.
}
\label{fig:roi}
\end{figure}

\subsection{Real-world Env: RoI-aware Exposure Control}
\quad Our DRL-AE framework can also take arbitrary input sizes because the framework resizes the image before the vectorized intensity processing.
Also, our domain randomization strategy produces a random size of the Region of Interest (RoI) patch by using crop, flip, resize, and rotation in the training stage.
Therefore, our agent is able to control camera exposure for a specific RoI or entire image.
\figref{fig:roi} shows the RoI-aware exposure control results.

The agent adjusts the camera parameters to maximize the image attribute for a specific RoI area.
It allows the camera to capture the detailed context for the regions of interest.
We expect that DRL-AE can be combined with object detection, object tracking, and human gaze and attention.
As a result, the combination can lead to adaptive exposure control schemes, such as attention-aware or detection-aware exposure control.

\subsection{Computation Time Analysis}
\quad Our agent has a simple Multi-Layer Perceptron (MLP) architecture with two hidden layers of 256 units. Also, our method does not require complex matrix computation like convolutions. 
Therefore, our agent can be run on a CPU device in real-time.
We measure the inference time of the agent on the Ryzen 5950x CPU. 
\tabref{tbl:calcuationtime} shows the computation time results, compared with shin \etal \cite{shin2019camera}.
The network's inference time takes 1 \textit{ms} regardless of image resolution because the input image is resized to a fixed resolution. 
Also, even including other operations, such as image resizing and RoI cropping, it takes a maximum 6 \textit{ms}, which is still in the real-time range.
Therefore, it can run at 170-1000 Hz on a CPU device.

\begin{table}[t]
\caption{\textbf{Processing time analysis.} Our algorithm does not use any complex metric or computation; the agent consists of two MLP layers with 256 hidden units, and the vectorized intensity history does not need complex operations. Therefore, our agent can be run on a CPU device in real-time. Here, we measure the processing time on the Ryzen 5950x CPU.}
\label{tbl:calcuationtime}
\begin{center}
\begin{small}
\begin{tabular}{c|cc}
\toprule
Method & Image Size & Processing Time (\textit{ms}) \\
\midrule
\multirow{2}{*}{Shin \etal \cite{shin2019camera}} & 1600 x 1200 & 108.7 \\
 & 800 x 600 & 18.2 \\
 \midrule
 \multirow{2}{*}{Ours} & 1600 x 1200 & 1.0 \\
 & 800 x 600 & 1.0 \\
 \bottomrule
\end{tabular}
\end{small}
\end{center}
\vspace{-0.1in}
\end{table}

\section{Conclusion \& Future Work}
\quad \textbf{Conclusion.}
In this paper, we proposed a novel joint exposure parameter control framework that exploits Deep Reinforcement Learning (DRL) to achieve instant exposure convergence and real-time processing.
The proposed framework, named DRL-AE, effectively solves the challenges when applying DRL to the exposure control task, such as 1) training environment to provide diverse lighting change scenarios, 2) flickering and image attribute-aware reward design, 3) lightweight state design by using vectorized intensity history, and 4) domain generalization via spatial domain randomization strategy.

The proposed method is thoroughly validated in three different environments: light-controlled dark room, exposure control dataset~\cite{shin2019camera}, and real-world environments.
We demonstrate that our proposed method instantly adjusts camera exposure within five steps with real-time processing of 1 \textit{ms} on a CPU device.
Also, our method shows satisfactory generalization performance in the wild. 
The images acquired from our method are well-exposed and show superiority in numerous computer vision tasks, such as feature extraction and object detection\footnote{Further visualization and video demos are available at \url{https://sites.google.com/view/drl-ae}.}.
To the best of our knowledge, our approach is the first solution that applies DRL to control camera exposure.
We hope our paper encourages active research of advanced camera exposure control algorithms to achieve robust visual perception ability.

\textbf{Future Work.}
This paper shows that DRL can be used in the field of camera exposure control.
There are lots of open research topics, such as motion-aware AE control, advanced reward function, aperture control, hardware generalization over various cameras, and further domain generalization in the real world.
In the future, we plan to extend the current darkroom environment to generate object or camera motion, allowing the agent to consider motion blur for exposure parameter control. 
Controlling camera aperture by using a mechanical aperture control module is another research direction.

\section*{Acknowledgment}
This research was supported by a grant (P0026022) from R\&D Program funded by Ministry of Trade, Industry and Energy of Korean government. 

\clearpage
\maketitlesupplementary

\section{Appendix}
In this supplementary material, we provide

\begin{itemize}
\item  Implementation and training details
\item  Further discussion of RL component design
\item  Discussion and future works
\end{itemize}

\section{Implementation Details}
\quad \textbf{Network Architecture.} 
We adopt simple Multi-Layer Perceptron (MLP) layers as our DRL agent architectures (\ie, actor and critic).
\tabref{tbl:architecture} shows the architecture detail from the input to the output layers.
Both actor and critic networks consist of one input layer, two intermediate layers, and one output layer.
The dimensions of each network's input and output layers are determined by the dimensions of state and action vectors.
Given the state vector, the actor network estimates next step actions (\ie, exposure time and gain difference). 
The critic network receives a state and action as input and estimates a q-value. 
This q-value is then utilized in the soft actor-critic training process \cite{haarnoja2018soft}.

\textbf{Training Settings}. 
We use a machine equipped with a Ryzen 5950x CPU and NVidia 3080Ti GPU for the agent training. 
We use Adam optimizer~\cite{kingma2014adam} with an initial learning rate of $3\cdot10^{-4}$. 
The agent is trained with a batch size of 256 for 500k timesteps in a light-controlled darkroom environment. 
We set a maximum exposure value of 100 $ms$ and a maximum gain of 40 $dB$ for the control bound of the machine vision camera. 
In the darkroom environment with controlled LED lighting, the agent stores various exposure transition sets in the replay buffer for each episode.
The actor and critic networks are optimized using the transition batches sampled from the buffer.
The maximum episode length is set to 200 steps. 
It usually takes about 20 seconds per episode, including training and image acquisition time. 
Also, the agent is validated every 2000 time steps. 
The total training time usually takes 18 hours in these training conditions. 
For the other hyperparameters, we follow the default setting described in~\cite{haarnoja2018soft} and summarize them in~\tabref{tbl:hyperparameters}. 

\begin{table}[h!]
  \caption{{\bf Network architectures details.} }
  \label{tbl:architecture}
  \centering
  \begin{adjustbox}{max width=0.95\linewidth}
  \begin{tabular}{l|cc}
    \toprule
    Layer Type& Actor &  Critic \\
    \hline
    Linear      & (state\_dim, 256)     &  (state\_dim + action\_dim , 256) \\
    Activation  & ReLU                  &   ReLU \\
    Linear      & (256, 256)            &   (256, 256) \\
    Activation  & ReLU                  &   ReLU \\
    Linear      & (256, 256)            &   (256, 256) \\
    Activation  & ReLU                  &   ReLU \\
    Linear      & (256, $2\cdot$action\_dim)    &   (256, 1) \\
    \bottomrule
  \end{tabular}
  \end{adjustbox}
\end{table}

\begin{table}[b]
  \caption{{\bf Hyperparamters for agent training.}}
  \label{tbl:hyperparameters}
  \centering
  \begin{adjustbox}{max width=0.95\linewidth}
  \begin{tabular}{l|c}
    \toprule
    Parameter & Value  \\
    \hline
    optimizer           & Adam \cite{kingma2014adam} \\
    learning rate       & $3\cdot10^{-4}$ \\
    discount ($\gamma$) & 0.99 \\
    replay buffer size  & $10^6$ \\
    target smoothing coefficient ($\tau$) & 0.05 \\
    batch size & 256 \\
    initial random steps & 10000 \\
    \bottomrule
  \end{tabular}
  \end{adjustbox}
\end{table}

\begin{figure}[t]
\begin{center}
\begin{tabular}{c}
\includegraphics[width=0.8\linewidth]{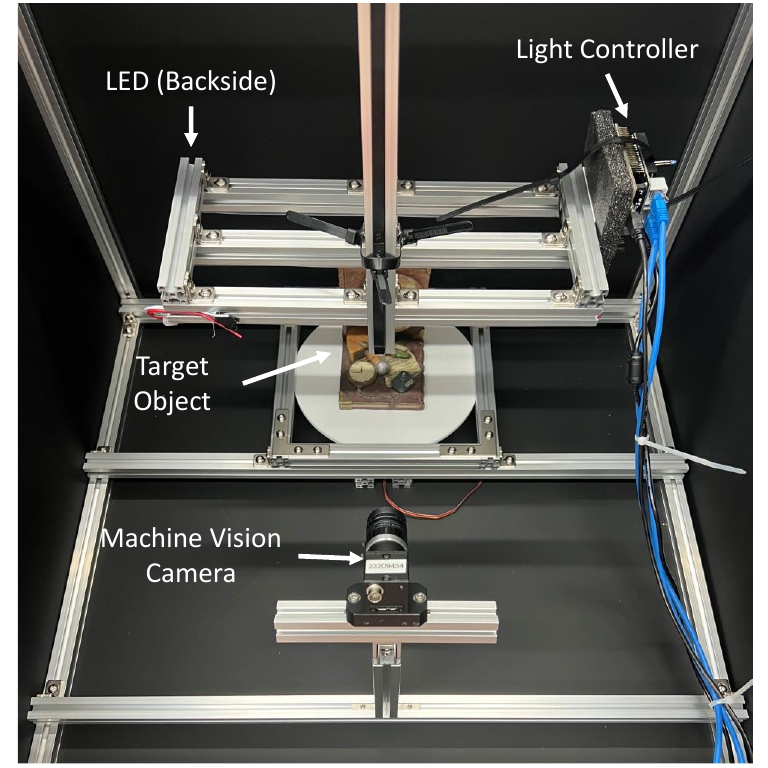} 
\end{tabular}
\end{center}
\vspace{-0.2in}
\caption{{\bf Light-controlled darkroom.} }
\label{fig:dark_room}
\end{figure}

\begin{figure}[t]
\begin{center}
\begin{tabular}{c}
\includegraphics[width=0.9\linewidth]{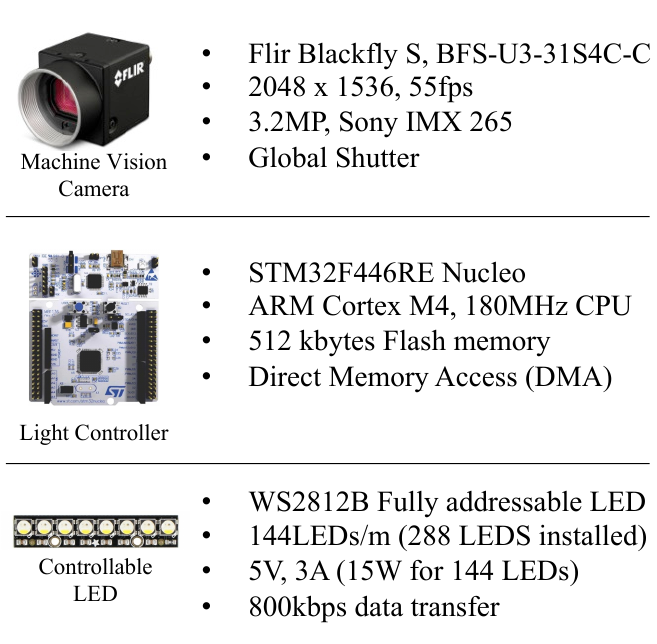} 
\end{tabular}
\end{center}
\vspace{-0.2in}
\caption{{\bf Hardware specification used in the darkroom.}}
\label{fig:sensorspec}
\end{figure}

\textbf{Light-controlled Darkroom}. 
We build the darkroom environment to freely control the lighting. 
Our aim is to provide a various range of lighting conditions to the RL agent, within a short training time compared to the real sunlight condition. 
The darkroom environment is made with aluminum profiles and black acrylic plates of 5 mm thickness. 
\figref{fig:dark_room} and \figref{fig:sensorspec} shows the constructed environment and each component's specification.
We use a global shutter machine vision camera from Teledyne FLIR, which has a 3.2MP Sony IMX 265. 
For the light controller unit, we built a program based on the STM32F446RE Necleo board, which has a 180MHz ARM Cortex M4 CPU and a flash memory of 512 kbytes.
Our LED bar is based on the WS2812B LEDs, which have 144 LEDs per meter. 
We use two LED bars for the darkroom environment. 
The light controller communicates with the RL gym environment located on the RL server through serial communication.

\section{Design Philosophy for RL Component}
In this section, we describe the underlying philosophy of designing the RL components for the exposure control task. 

\subsection{State Design}

\quad \textbf{CNN Feature.} 
Perhaps, a na\"ive state designing is utilizing the CNN model to extract a feature map from the image and use the feature map as a state.
However, CNN-based state design has three disadvantages.
First, the extracted feature map has no clear relationship with the camera exposure level. 
Usually, CNN backbones (\eg, ImageNet, VGG, ResNet) are trained in a brightness-agnostic manner via their data augmentation strategy.
Therefore, the extracted feature usually includes semantic information rather than brightness information.
Second, CNN brings additional domain gap problems in real-world inference.
Third, the state extraction with the CNN model is computationally heavy, introduces additional learnable parameters, and requires lots of system memory to store the image states in the replay buffer.
As a result, CNN state brings undesirable properties for deep reinforcement learning, such as reducing replay buffer size, limiting sample diversity, increasing training time, generalization problems, and unclear representation of brightness.

\textbf{Intensity Value.} 
Therefore, instead of using the CNN model, we utilize the averaged intensity values along the x-axis.
The primary purpose of auto-exposure control is to ensure a proper image brightness level by adjusting camera exposure settings.
Therefore, image intensity is the primary cue and a straightforward and effective representation for auto-exposure control.
Also, we reduce the dimension by averaging intensity value along the x-axis rather than utilizing entire images to ensure high-level memory efficiency with minimum computation burden.
Lastly, we stack 1-dimensional averaged intensity values of 3 frames and define the stacked intensity values as a \textit{state}.
We empirically found that frame stacking has a significant impact on making better decisions. 
By stacking consecutive frames, the agent is able to implicitly observe the lighting condition change and camera exposure change over a short period.

\subsection{Action Design}
\quad \textbf{Discrete \textit{vs.} Continuous Action Space.} 
We initially considered designing the system to obtain quantized exposure and gain values using a discrete action space. 
However, due to the sensitive nature of the parameter optimization, which often causes oscillation, it was difficult to fully account for the changes even with a higher level of quantization. 
Therefore, we formulate the auto exposure control problem as a continuous action control task.

\textbf{Absolute \textit{vs.} Relative Action Range.}
We can consider absolute and relative actions as output values.
The former indicates the agent estimate desired absolute action values (\ie, absolute values of exposure time and gain, such as 10 $ms$, 4$dB$).
The latter means the output is a relative difference in action values (\ie, $\pm$10$ms$, $\pm$2$dB$)
Relative control is more stable but has the disadvantage of slower convergence. 
On the other hand, absolute control can reach the desired value in one step but is more likely to be unstable. 
In practice, we found that absolute control faced difficulties in learning good policies.
Therefore, we make the agent estimate relative action for the exposure control task.

\subsection{Reward Design}
\quad Designing reward functions is crucial in deep reinforcement learning, as it determines the desired objective of the learning process.
In this context, we will briefly mention the rationale behind the three reward functions presented in the main text.
A common desired goal for auto exposure control is to acquire a high-quality image that has moderate brightness, low-level noise, and sharp edge information. 
Also, the convergence of the control process should be fast but stable. 

Therefore, the proposed reward functions are designed for these desired objectives.
First, the mean reward term $\mathcal{R}_{mean}$ helps to ensure that the image has a median brightness. 
Instead of designing it linearly, we made it decrease more steeply around the median by adding non-linearity with $p_m$ to maintain the center.
The second flickering term $\mathcal{R}_{flk}$ suppresses the image flickering effect caused by the action's vibration and ensures smooth exposure transition while preserving image attributes.
Lastly, the noise term $\mathcal{R}_{noise}$ reduces the overall image noise caused by excessively high gain, encouraging a balanced control between the exposure and gain parameters.

\section{Discussion and Future Work}
\begin{figure}[t]
\begin{center}
\begin{tabular}{c@{\hskip 0.005\linewidth}c@{\hskip 0.005\linewidth}c}
\includegraphics[width=0.33\linewidth]{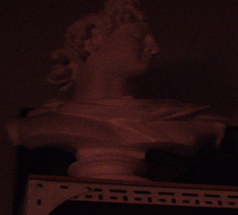} &
\includegraphics[width=0.33\linewidth]{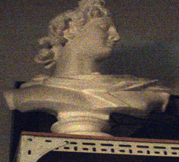} &
\includegraphics[width=0.33\linewidth]{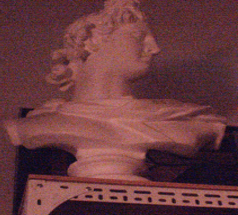} \\
{\footnotesize (a) Built-in AE} & {\footnotesize (b) w/ Contrast\&Tone} & {\footnotesize (c) w/ Zero-DCE~\cite{guo2020zero}} \\
\includegraphics[width=0.33\linewidth]{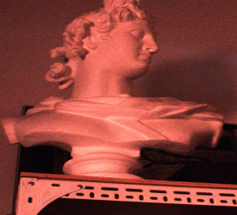} &
\includegraphics[width=0.33\linewidth]{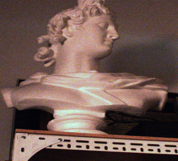} &
\includegraphics[width=0.33\linewidth]{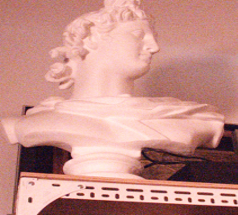} \\
{\footnotesize (d) Ours} & {\footnotesize (e) w/ Contrast\&Tone} & {\footnotesize (f) w/ Zero-DCE~\cite{guo2020zero}} \\
\end{tabular}
\end{center}
\vspace{-0.1in}
\caption{{\bf Impact of AE control on post-processing methods.}}
\label{fig:rebuttal_enhance}
\end{figure}

\quad \textbf{Camera AE vs. ISP.}
There are two main processes for camera image processing: Automatic Exposure (AE) control and Image Signal Processing (ISP).
AE control includes automatic gain, exposure, and aperture control as well.
The roles of AE control and ISP are quite different.
The former aims to get high-quality images while rapidly adjusting exposure levels within hardware limitations.
After that, the latter enhances the quality of the acquired image for its purpose by using various ISP tools, such as demosaicking, deblurring, denoising, color space correction, gamma correction, tone-mapping, HDR, and more.
Therefore, the AE control (hardware level) and ISP (software level) are complementary rather than competitive relations. 
As the former stage obtains a better quality image, the quality in the later stage improves.
We evaluate our method by combining conventional contrast enhancement \& tone mapping method (\ie, photoshop) (e) and Zero-DCE~\cite{guo2020zero} (f).
As shown in~\figref{fig:rebuttal_enhance}, the exposure control results highly affect its final outputs ((b)vs(e), (c)vs(f)).

\begin{figure}[t]
\begin{center}
\begin{tabular}{cc}
\includegraphics[width=0.4\linewidth]{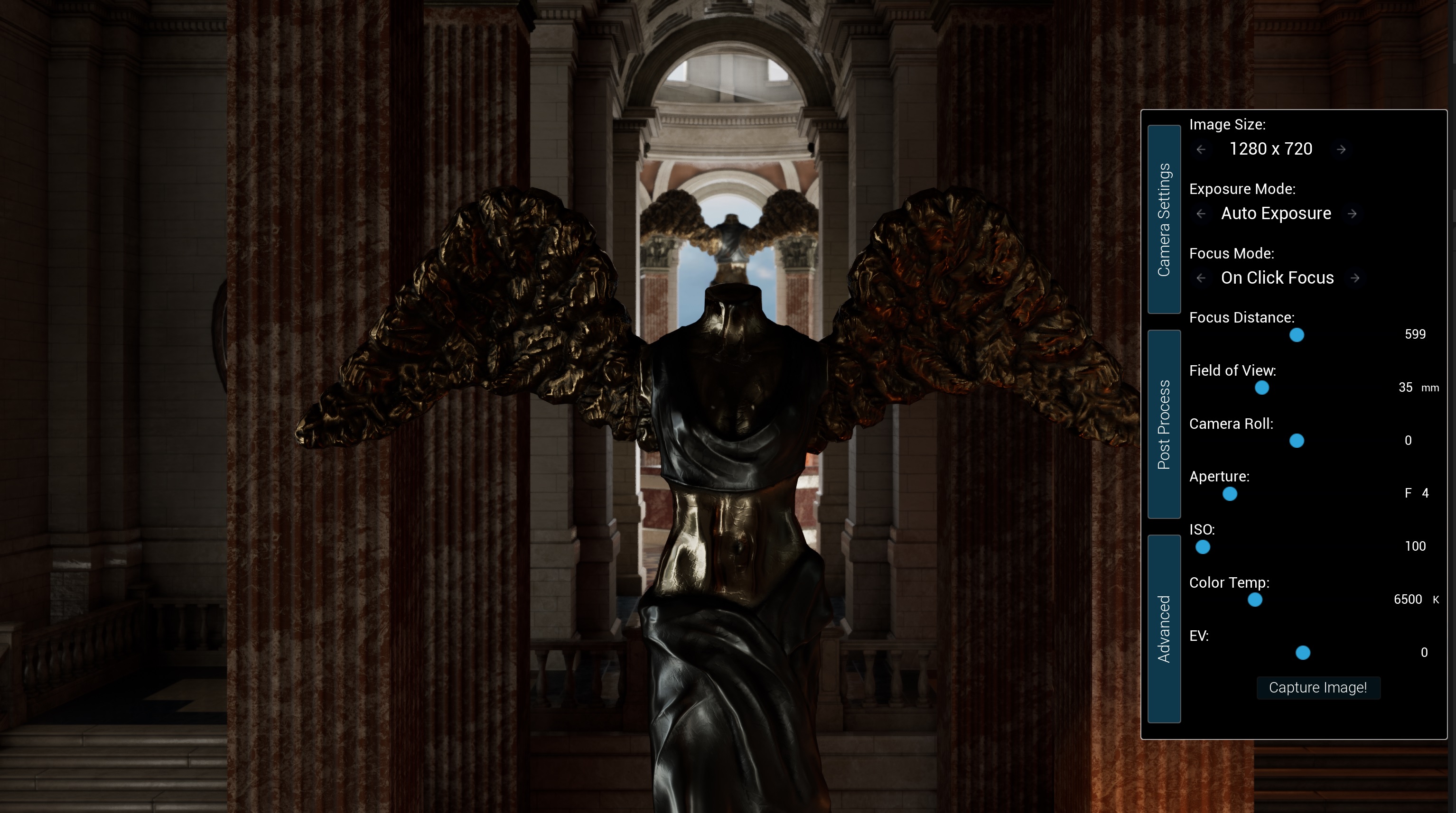} &
\includegraphics[width=0.4\linewidth]{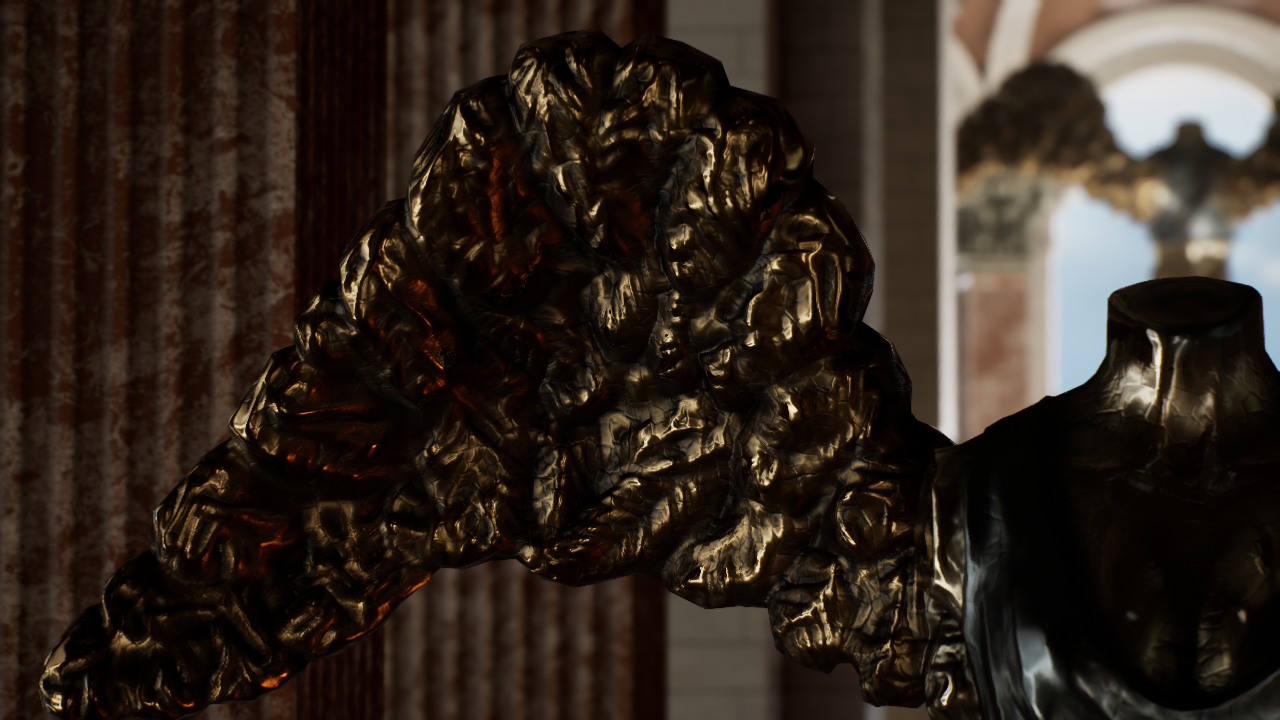} \\
{\footnotesize(a)} & {\footnotesize(b)}\\
\includegraphics[width=0.4\linewidth]{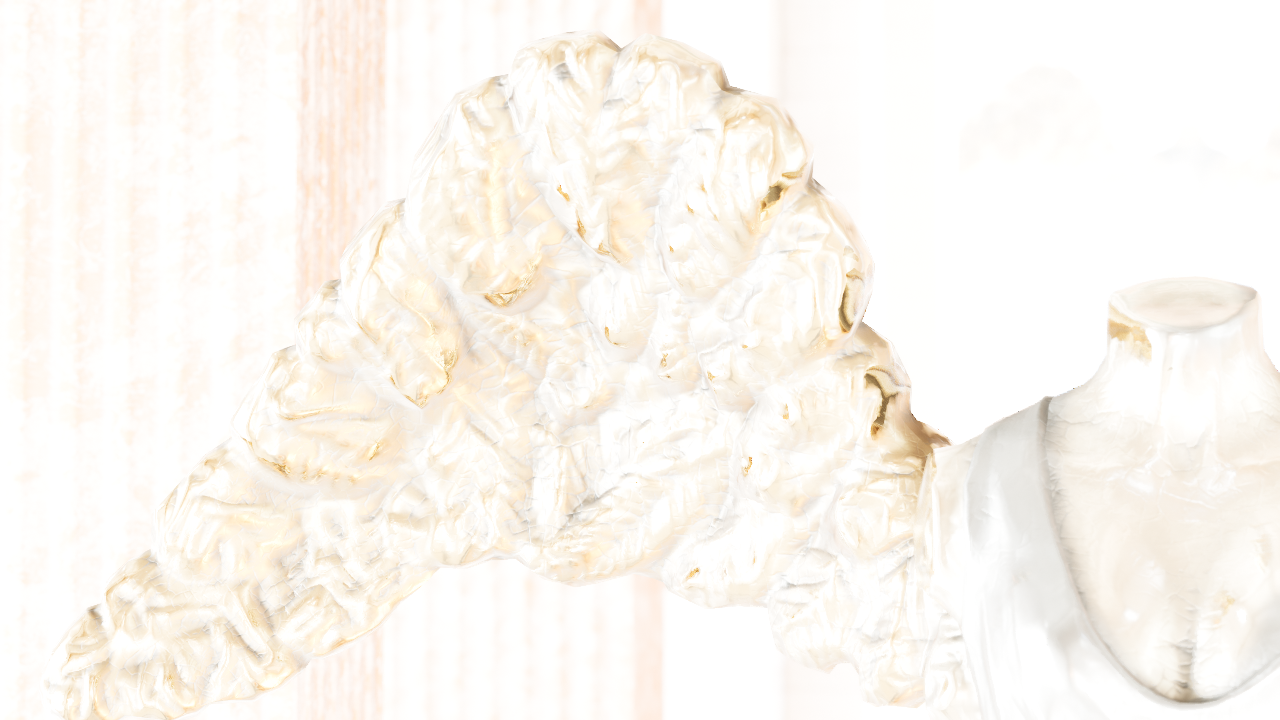} &
\includegraphics[width=0.4\linewidth]{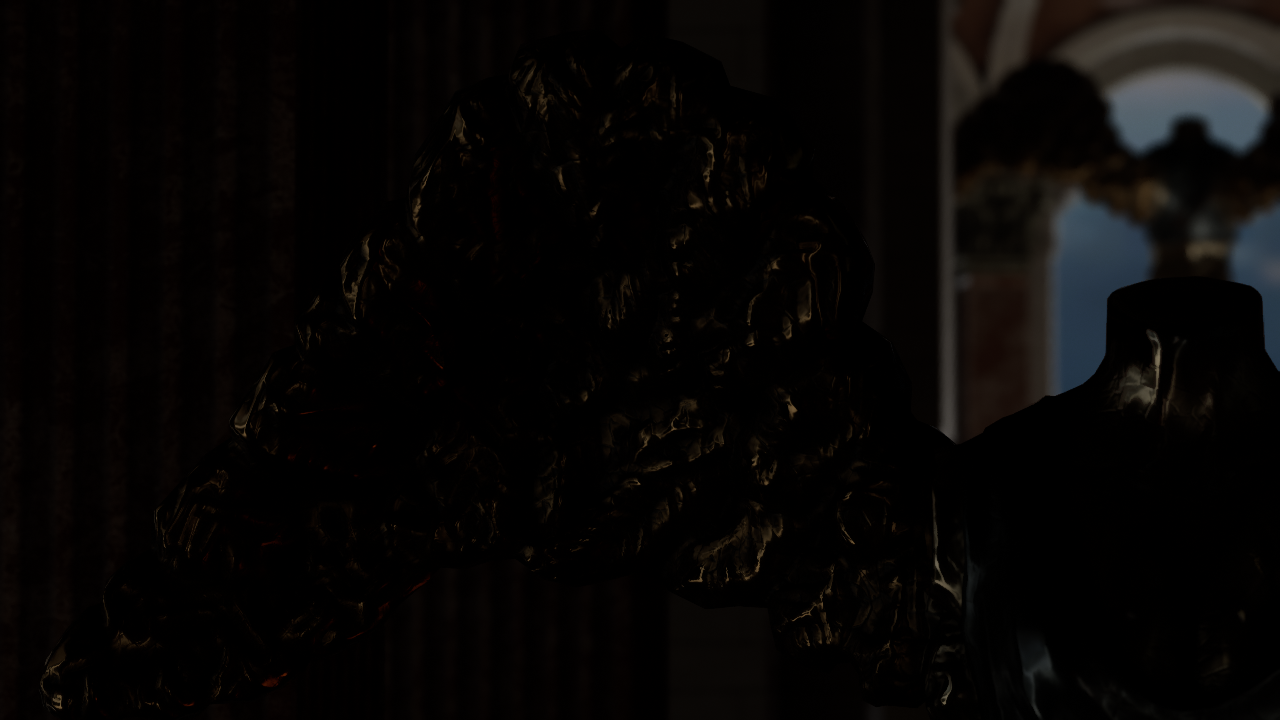} \\
{\footnotesize(c)} & {\footnotesize(d)}\\
\end{tabular}
\caption{
{\bf Example image from Unreal-based camera simulator \cite{unrealsym}.} (a) Interface overview, (b) Auto-Exposure, (c) Over-exposed, (d) Under-exposed. }
\label{fig:simulators}
\end{center}
\vspace{-0.3in} 
\end{figure}

\textbf{Sim2Real via Camera Simulator. }
From the perspective of object, motion, and lighting diversity, a simulated camera model could be beneficial to learning camera exposure control with deep reinforcement learning.
Modern photorealistic simulations, such as Unreal, Unity, and Blender, provide similar quality images to the real world and support partial functionality for auto-exposure control.

However, introducing simulation causes two domain gap issues: the domain gap 1) between actual and simulated environments and 2) between simulated and real camera acquisition models.
The former issue leads to large performance differences between the models trained on simulated data and real data, as we can see in domain adaptation literature.
For the latter issue, the simulated camera model provides an \textit{incomplete} image acquisition model.
As shown in~\figref{fig:simulators}-(c), when the camera captures the image with high gain or ISO, the image must contain severe noise within the image. 
However, the simulation doesn't support this functionality.
Therefore, due to these issues, we decided to utilize the darkroom environment to investigate the possibility of deep reinforcement learning for automatic exposure control.

\textbf{Future Work: DRL-AE in Simulation.} 
Although the simulation has some disadvantages, it has many advantages, such as faster interaction speed, easy parallelization, and diverse controllable parameters.
Therefore, in future work, we also plan to study Sim2Real based camera exposure control and compare the Sim2Real model with the real-world model trained with this paper's method.

\textbf{Future Work: Motion-aware AE Control.} 
Considering the motion blur is another future direction. 
As the proposed darkroom environment has only a fixed target object, it is difficult to consider motion blur that frequently happens in the real world.
In future work, we plan to extend the current darkroom environment to make object motion, thus allowing the agent to consider a motion blur for their exposure parameter control.

\textbf{Future Work: Various Reward Functions.}
In this paper, the proposed reward design might be a primitive and basic form for camera exposure control.
However, it can be easily extendable by incorporating modern image assessment metrics~\cite{shin2019camera,kim2020proactive,han2023camera}. 
Also, we can utilize human preference, network inference results (\eg detection confidence), and the number of detected features as a reward function.

\textbf{Future Work: Aperture Control.}
The machine vision camera used in the experiment has a fixed aperture size, which is not controllable with software.
However, the aperture is also one parameter that affects camera exposure level and depth of field.
Therefore, we plan to control aperture size by using a mechanic aperture control module.

{
    \small
    \bibliographystyle{ieeenat_fullname}
    \bibliography{egbib}
}

\end{document}